\documentclass[11pt]{article}

\usepackage[final]{acl}
\usepackage{times}
\usepackage{latexsym}
\usepackage[T1]{fontenc}
\usepackage[utf8]{inputenc}
\usepackage{microtype}
\usepackage{inconsolata}
\usepackage{graphicx}
\usepackage{booktabs}
\usepackage{array}
\usepackage{amsmath}
\usepackage{amssymb}
\usepackage{multirow}
\usepackage{enumitem}
\usepackage[table]{xcolor}

\newcommand{\method}{\textsc{PRSR}}

\newcommand{\hgain}[1]{\textcolor[RGB]{210, 50, 45}{$_{\mathbf{\uparrow #1}}$}}

\title{Steering beyond Language:\\ Improving Multilingual Training-Free Text Embeddings via\\ Paired-Residual Subspace Removal}

\author{
    Ailiang Lin$^{1}$, 
    Zhuoyun Li$^{2}$,
    Yusong Wang$^{1}$,
    Keyu Mao$^{1}$,\\
    \textbf{Kotaro Funakoshi}$^{1}$,
    \textbf{Manabu Okumura}$^{1}$\\
    $^{1}$Institute of Science Tokyo \quad $^{2}$Tencent\\
\texttt{\{linailiang, wangyi, maokeyu, funakoshi, oku\}@lr.first.iir.isct.ac.jp}\\ \texttt{earyli@tencent.com}
}

\begin{document}
\maketitle

\begin{abstract}
While deriving text embeddings directly from frozen LLMs has emerged as a promising direction, existing methods focus almost exclusively on English, leaving multilingual settings largely unexplored. In this work, we present the first systematic study of multilingual training-free text embeddings in LLMs. Our analysis reveals a language-conditioned embedding geometry in which semantically equivalent sentences remain misaligned across languages. We hypothesize that language-specific variation can be disentangled from semantic content to improve embedding quality. To this end, we propose \textit{Paired-Residual Subspace Removal} (PRSR), a training-free inference-time intervention that steers LLM representations toward a more language-invariant semantic space. PRSR first constructs paired residuals from multilingual parallel sentences to capture language-specific deviations. It then estimates layer-wise low-rank subspaces from these residuals and suppresses the corresponding components in the last-token hidden states. We conduct more than 200 monolingual evaluations across 18 languages using MTEB and MMTEB, along with cross-lingual bitext mining between English and over 200 other languages. Empirical results demonstrate that PRSR consistently improves the performance of existing training-free baselines in both monolingual and cross-lingual settings across different LLMs.
\end{abstract}

\begin{figure}[t]
\centering
\includegraphics[width=\columnwidth]{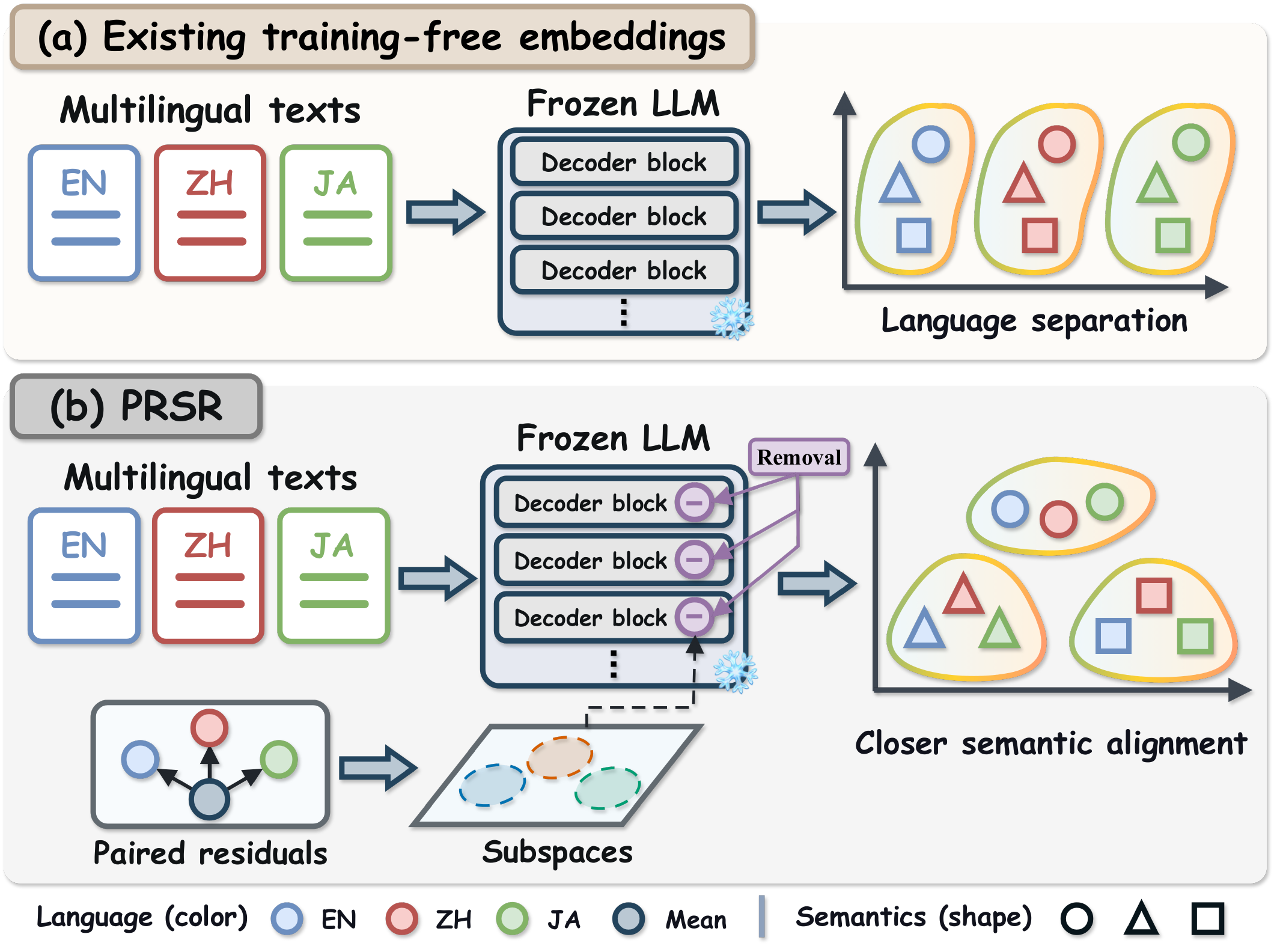}
\caption{Illustration of \method{}. (a) Training-free text embeddings remain separated by languages despite shared semantics. (b) \method{} estimates language-sensitive subspaces from paired residuals of parallel sentences and removes the corresponding components within the frozen LLM to improve semantic alignment.}
%Colors denote languages; shapes denote meanings.
\label{fig:teaser}
\end{figure}

\section{Introduction}
Text embeddings underpin applications such as information retrieval, recommendation, and clustering~\cite{mteb,mmteb}. While large language models (LLMs) exhibit remarkable semantic understanding, transforming them into embedding models typically requires extensive fine-tuning~\cite{llm2vec,lin-etal-2026-embedding}. Training-free methods offer an appealing alternative by deriving embeddings directly from frozen LLMs. PromptEOL~\cite{prompteol} pioneers this direction by prompting LLMs to condense sentence semantics into a single word. Subsequent work has advanced this paradigm through prompt design~\cite{metaeol,echo} or internal model interventions~\cite{tp,cp,kvembedding,htp}.

However, existing training-free methods have been developed and evaluated predominantly in English, leaving their multilingual behavior largely unexamined. We address this gap through the first systematic study of training-free text embeddings in LLMs across languages. Our motivating analysis of multilingual parallel texts shows pronounced clustering by language: translations conveying the same meaning tend to occupy distinct regions of the last-token latent space. Crucially, semantic equivalence across languages is not consistently reflected in embedding similarity. These findings reveal a language-conditioned embedding geometry in which language identity can obscure semantic relationships between sentences.

Notably, prior work suggests that language-specific information can be localized within LLMs~\citep{tang-etal-2024-language,zhao2024how}. Manipulating such language-related directions has further been shown to improve multilingual reasoning~\citep{zhao2025when,maraia-etal-2026-activation}. This raises a natural question for training-free text embeddings: \textit{can language-specific variation be suppressed without sacrificing semantics?}

To this end, we propose \textbf{PRSR} (\textbf{P}aired-\textbf{R}esidual \textbf{S}ubspace \textbf{R}emoval), an inference-time activation steering method for improving multilingual training-free text embeddings in LLMs. \method{} exploits the semantic correspondence of multilingual parallel sentences to separate language-specific variation from shared semantic content. Specifically, at each selected layer, \method{} constructs \textit{paired residuals} by subtracting the group mean from the representation of each parallel sentence. The dominant singular directions of the stacked residual matrix define a low-rank language-sensitive subspace. At inference time, \method{} removes the projections of last-token hidden states onto these subspaces, allowing the modified representations to propagate through the remaining frozen layers and influence later computation. As a result, \method{} produces text embeddings that are less sensitive to language identity yet more closely aligned by semantic similarity, improving semantic discrimination within languages and alignment across languages. This design can be readily integrated with existing training-free embedding methods without model parameter updates.

We conduct extensive experiments on MTEB and MMTEB~\cite{mteb,mmteb}, comprising more than 200 monolingual evaluations in 18 languages. In addition, we evaluate cross-lingual bitext mining on BUCC.v2~\cite{bucc}, Tatoeba~\cite{artetxe-schwenk-2019-massively}, and FLORES~\cite{flores1,flores2} between English and over 200 other languages. Experimental results across three LLMs show that \method{} consistently improves existing training-free baselines in both monolingual and cross-lingual settings. Comprehensive ablation studies and analyses further validate the effectiveness of the proposed steering mechanism. Overall, this work highlights that suppressing language-specific variation within frozen LLMs can improve multilingual text embedding quality. 

The main contributions of this work are summarized below: (1) We present the first systematic study of multilingual training-free text embeddings in LLMs. Our motivating analysis reveals a language-conditioned embedding geometry in which language identity can interfere with semantic relationships. (2) We propose \method{}, a training-free inference-time steering method that constructs paired residuals from multilingual parallel sentences to estimate language-sensitive subspaces and removes the corresponding components during embedding extraction. (3) Comprehensive experiments on monolingual MTEB/MMTEB evaluations and cross-lingual bitext mining benchmarks demonstrate that \method{} consistently improves existing training-free baselines across various LLMs.

\section{Related Work}
\paragraph{Training-Free Text Embeddings.}
Recent advances in LLMs have spurred growing efforts to adapt them into dedicated embedding models \citep{geminiembedding,qwen3embedding}. However, such approaches require substantial training data and fine-tuning. Training-free methods offer a practical alternative by extracting text embeddings directly from frozen LLMs through prompt design or internal model interventions. PromptEOL~\citep{prompteol} uses a one-word constraint to condense sentence semantics into the last-token representation, while Pretended CoT~\citep{cot} augments the prompt with a chain-of-thought instruction. ECHO~\cite{echo} repeats the input so that tokens in the second occurrence can attend to the full sentence, while ReverseEOL~\citep{reverseeol} combines embeddings of the original and reversed text to capture complementary context. Another line of work explores interventions on internal representations. Contrastive Prompting~\cite{cp} steers hidden states by contrasting target and auxiliary prompt activations, Token Prepending~\cite{tp} prepends each layer’s last-token representation to the next layer’s input, and KV-Embedding~\citep{kvembedding} reroutes last-token key--value states as internal prefixes for global context access. These methods have been evaluated primarily in English, leaving their effectiveness in multilingual settings largely unexplored. Our work addresses this gap by systematically examining their behavior across languages.

\paragraph{Language-Agnostic Representations.}  
Multilingual embedding models commonly learn cross-lingual alignment from parallel data~\citep{artetxe-schwenk-2019-massively,feng-etal-2022-language}. To further reduce language-specific information, prior work suppresses language signals during retriever training~\citep{huang-etal-2024-language} or factorizes multilingual representation learning into specialized and shared components~\citep{huang-etal-2025-modular}. Alternatively, LIR~\citep{yang-etal-2021-simple} and LSAR~\cite{xie-etal-2022-discovering} remove language-associated subspaces from extracted embeddings without additional training. In contrast, \method{} modifies intermediate hidden states within the frozen LLM, allowing the intervention to influence subsequent computation.

\paragraph{Activation Steering in LLMs.}
Activation steering guides model behavior through inference-time modifications to intermediate activations rather than parameter updates. For multilingual processing, prior studies have identified language-specific neurons and functional specialization across LLM layers~\citep{tang-etal-2024-language,zhao2024how}. \citet{dumas-etal-2025-separating} suggest that language and conceptual information can be manipulated independently within the internal representations of LLMs. Recent studies further explore multilingual activation steering: language--reasoning disentanglement improves multilingual reasoning by suppressing language-specific activations \citep{zhao2025when}, while steering directions learned in English generalize to syllogistic reasoning in other languages \citep{maraia-etal-2026-activation}. We extend multilingual activation steering to text embedding generation, improving semantic representations in both monolingual and cross-lingual settings.

\section{Multilingual Analysis}
\label{sec:observation}
In this section, we investigate whether representations preserve semantic correspondence across languages. We elicit training-free text embeddings from a frozen LLaMA-2-7B~\citep{llama2} using PromptEOL~\citep{prompteol}. On FLORES+ parallel texts~\citep{flores1}, we first characterize language-specific clustering and then examine whether the shared semantics of parallel texts are reflected in their embedding similarity.

\subsection{Language-Conditioned Embedding Geometry}
We use t-SNE to visualize text embeddings from $1500$ FLORES+ parallel sentence groups across eight languages. Each group contains translations of the same sentence, ensuring that they share the same underlying semantics.

\begin{figure}[t]
\centering
\includegraphics[width=\columnwidth]{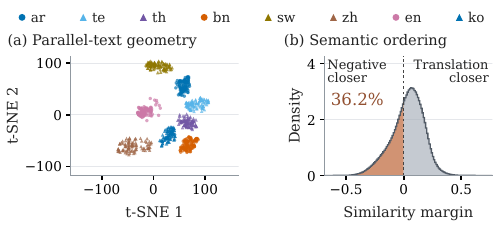}
\caption{Language-conditioned geometry of training-free text embeddings on FLORES+. (a) t-SNE visualization showing language-specific clustering. (b) Cosine-similarity margins between anchor--translation and anchor--negative pairs; in 36.2\% of comparisons, negatives are more similar to anchors than translations.}
\label{fig:preliminary-findings}
\end{figure}

Figure~\ref{fig:preliminary-findings}(a) shows pronounced clustering by language: sentences with diverse meanings occupy language-specific regions, whereas corresponding translations remain separated. Given the matched semantic content, this pattern suggests a language-conditioned embedding geometry in which language identity is a prominent organizing factor, consistent with prior findings that multilingual embeddings retain language-specific information alongside shared semantics~\citep{xie-etal-2022-discovering,huang-etal-2024-language}.

\begin{figure*}[t]
\centering
\includegraphics[width=\textwidth]{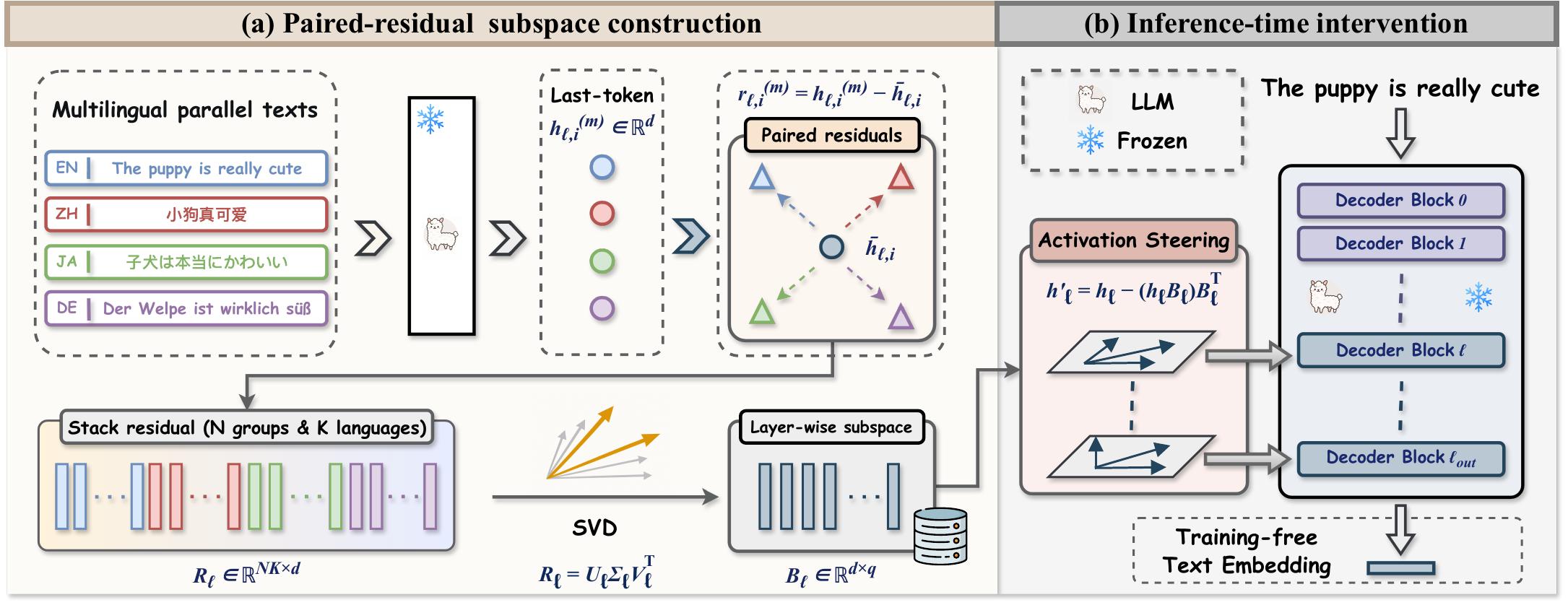}
\caption{Overview of \method{}. \textbf{(a) Paired-residual subspace construction.} At each selected layer, we first construct paired residuals by subtracting the group mean from the last-token representations of parallel sentences. We then stack the residuals across groups and languages, and apply SVD to estimate a low-rank language-sensitive subspace. \textbf{(b) Inference-time intervention.} We remove the projections of last-token hidden states onto these subspaces at the corresponding layers of the frozen LLM, steering representations toward a more language-invariant semantic space.}
\label{fig:method}
\end{figure*}

\subsection{Language Variation Rivals Semantic Variation}
\label{sec:semantic-displacement}
Parallel texts allow us to further assess whether similarity between cross-lingual embeddings reliably reflects shared meaning across languages. For each anchor sentence, we compare two cosine similarity scores: one for its \textit{translation}, which preserves semantics across languages, and the other for a \textit{negative} sentence that matches the anchor in language identity but differs in semantic content. We use $1{,}500$ FLORES+ parallel groups spanning 18 languages and select negatives through length-stratified sampling. Figure~\ref{fig:preliminary-findings}(b) shows the distribution of similarity margins, with positive values indicating higher similarity to the translation.

In 36.2\% of comparisons, the same-language negative is more similar to the anchor than the corresponding translation. This reveals that semantic equivalence across languages does not consistently yield a similarity advantage. In addition, the mean anchor--translation cosine distance is 85.6\% of the mean anchor--negative distance. Despite shared meaning, translations are nearly as distant in the embedding space as same-language sentences with different content. Together with the clustering results, these findings suggest that training-free text embeddings exhibit a language-conditioned geometry in which language identity can interfere with the underlying semantic structure. Additional analysis is provided in Appendix~\ref{sec:prsr-findings-comparison}.

\section{Paired-Residual Subspace Removal}
Existing training-free methods have been evaluated primarily on English embedding tasks, overlooking the potential impact of language-conditioned embedding geometry in multilingual settings. To this end, we propose \textbf{P}aired-\textbf{R}esidual \textbf{S}ubspace \textbf{R}emoval (\textbf{\method{}}), an inference-time steering method that suppresses language-specific variation while preserving shared semantics, as illustrated in Figure~\ref{fig:method}.

\subsection{Preliminaries}
Training-free text embeddings are typically read out from the last token in decoder-only LLMs, as it is the only position that can attend to the full sequence under causal attention. Formally, given an input text $x$, an embedding prompt forms the token sequence $p(x)=(u_1,\ldots,u_T)$. Let $h_{\ell}(x)\in\mathbb{R}^{1\times d}$ denote the hidden state of the last token after decoder block $\ell$, where $d$ is the embedding dimension. The text embedding is then extracted from the output layer as:
\begin{equation}
    e(x)=g\bigl(h_{\ell_{\mathrm{out}}}(x)\bigr),
    \label{eq:embedding-extraction}
\end{equation}
where $g$ denotes the model's output normalization.

\subsection{Paired-Residual Subspace Estimation}
The core idea of \method{} is to capture language-sensitive variation while controlling sentence semantics. To this end, we introduce \textbf{\textit{paired residuals}}, which measure how each translation deviates from the mean representation of its parallel group. Since all sentences within a group express the same underlying meaning, these residuals are expected to primarily capture language-related variation rather than semantic differences.

Given $N$ parallel groups in $K$ languages, let $x_i^{(m)}$ denote the sentence from group $i$ in language $m$, with last-token hidden state $h_{\ell,i}^{(m)}$ after decoder block $\ell$. We compute the group mean and define the corresponding paired residual $r_{\ell,i}^{(m)}$ as
\begin{equation}
%    \begin{aligned}
    \bar{h}_{\ell,i}
    =\frac{1}{K}\sum_{m=1}^{K}h_{\ell,i}^{(m)}, \ \ 
    r_{\ell,i}^{(m)}
    =h_{\ell,i}^{(m)}-\bar{h}_{\ell,i}.
%    \end{aligned}
    \label{eq:residual}
\end{equation}

We stack the paired residuals across all groups and languages into
$R_\ell\in\mathbb{R}^{NK\times d}$.
To identify the dominant directions of language-sensitive variation, we seek a $q$-dimensional subspace that minimizes the reconstruction error:
\begin{equation}
    B_\ell =
    \underset{B^\top B=I_q}{\arg\min}
    \left\|R_\ell-R_\ell BB^\top\right\|_F^2,
    \label{eq:subspace-objective}
\end{equation}
where $B\in\mathbb{R}^{d\times q}$ is an orthonormal basis. This objective identifies the subspace that captures the maximum variance of the paired residuals. Applying singular value decomposition (SVD) to $R_\ell=U_\ell\Sigma_\ell V_\ell^\top$ yields the optimal basis, consisting of the $q$ right singular vectors corresponding to the largest singular values:
\begin{equation}
    B_\ell=[v_{\ell,1},\ldots,v_{\ell,q}]
    \in\mathbb{R}^{d\times q},
    \label{eq:steering-basis}
\end{equation}
where $v_{\ell,j}$ denotes the $j$-th column of $V_\ell$. In practice, we estimate a separate subspace basis for each selected LLM layer and reuse these bases during evaluation across languages.

\subsection{Layer-Specific Activation Steering}
\label{sec:activation-steering}

Given the language-sensitive subspaces, \method{} steers last-token hidden states away from these subspaces during embedding extraction. Specifically, the projection onto $B_\ell$ corresponds to the dominant language-sensitive variation captured by the paired residuals. Removing this projection enables reducing language-related information, yielding more language-invariant semantic representations.

Starting from a designated block $\ell_s$, we apply the intervention at every decoder block up to the output layer $\ell_{\mathrm{out}}$, using the corresponding basis $B_\ell$. At each layer $\ell_s\leq\ell\leq\ell_{\mathrm{out}}$, we update the last-token hidden state $h_\ell(x)$ as
\begin{equation}
    \tilde{h}_\ell(x)
    =
    h_\ell(x)
    -
    \bigl(h_\ell(x)B_\ell\bigr)B_\ell^\top,
    \label{eq:intervention}
\end{equation}
where $\bigl(h_\ell(x)B_\ell\bigr)B_\ell^\top$ is the orthogonal projection of $h_\ell(x)$ onto the estimated language-sensitive subspace. After the final intervention, the text embedding is obtained as
\begin{equation}
    e_{\method{}}(x)
    =
    g\bigl(\tilde{h}_{\ell_{\mathrm{out}}}(x)\bigr).
\end{equation}
Since the intervention operates entirely on internal representations while keeping the underlying LLM frozen, \method{} can be readily integrated with existing training-free embedding methods to improve their performance in multilingual settings.

\section{Experiments}
\subsection{Experimental Setup}
\label{main:setup}
\paragraph{Benchmarks.}
We evaluate \method{} in both monolingual and cross-lingual settings. For \textit{monolingual evaluation}, we conduct $208$ task--language evaluations, spanning $18$ languages from MTEB~\citep{mteb} and MMTEB~\citep{mmteb}. In particular, we use MTEB (eng, v2), comprising $41$ datasets across seven embedding categories, for ablation and analysis. We report the average score for each language, together with an overall mean across all $208$ evaluations. More details on the dataset composition for each language are provided in Appendix~\ref{sec:monolingual-benchmarks}.

For \textit{cross-lingual evaluation}, we perform bitext mining on BUCC.v2~\citep{bucc}, Tatoeba~\citep{artetxe-schwenk-2019-massively}, and FLORES~\citep{flores1,flores2}, covering English paired with over $200$ other languages. For BUCC.v2 and Tatoeba, we report the mean score across all language pairs using each benchmark's primary metric. For FLORES, we report English-to-X and X-to-English scores separately, together with their mean. Additional benchmark details are provided in Appendix~\ref{sec:crosslingual-benchmarks}.

\paragraph{Baselines.}
We consider two families of training-free text embedding baselines.(i) \textit{Prompt-based methods.} \textsc{PromptEOL}~\citep{prompteol} extracts text embeddings using a one-word constraint. \textsc{Pretended CoT}~\citep{cot} extends PromptEOL with a chain-of-thought instruction. \textsc{ECHO}~\citep{echo} repeats the input and pools representations from the second occurrence. (ii) \textit{Internal model interventions.} \textsc{Contrastive Prompting (CP)}~\citep{cp} steers intermediate hidden states by contrasting target and auxiliary prompt activations. \textsc{Token Prepending (TP)}~\citep{tp} prepends each layer's last token to the input of the next layer.

\paragraph{Implementation.}
We conduct experiments and ablation studies using LLaMA-2-7B~\cite{llama2}, Mistral-7B~\cite{mistral}, and Qwen3-8B~\cite{qwen3}. All LLM parameters remain frozen, and text embeddings are extracted from the final output layer. To estimate the paired-residual subspaces, we use $1{,}500$ parallel sentence groups covering $K=15$ languages from the XNLI validation split~\citep{xnli}. For each LLM and training-free baseline, we estimate a separate set of layer-specific subspace bases, which are shared across all evaluation languages. Based on the analyses in Sections~\ref{abl:rank} and~\ref{abl:layer}, we set $q=K-1$ and $\ell_s$ to $17$, $14$, and $33$ for LLaMA-2-7B, Mistral-7B, and Qwen3-8B, respectively. All experiments are conducted on a single NVIDIA H100 96GB GPU. Further implementation details are provided in Appendix~\ref{sec:implementation-details}.

\begin{table*}[t]
  \centering
  \resizebox{\textwidth}{!}{
  \tabcolsep=4pt
  \begin{tabular}{l *{19}{c}}
  \toprule
  \textbf{Method} & \textbf{en} & \textbf{zh} & \textbf{fr} & \textbf{ru} & \textbf{de} & \textbf{ja}$^\dagger$ & \textbf{it}$^\dagger$ & \textbf{es} & \textbf{pt}$^\dagger$ & \textbf{ar} & \textbf{id}$^\dagger$ & \textbf{sw} & \textbf{vi} & \textbf{ko}$^\dagger$ & \textbf{bn}$^\dagger$ & \textbf{th} & \textbf{te}$^\dagger$ & \textbf{ms}$^\dagger$ & \textbf{AVG} \\
  \midrule
  PromptEOL & 46.8 & 38.7 & 43.4 & 45.0 & 39.4 & 46.6 & 55.1 & 47.9 & 47.9 & 41.2 & 49.6 & 42.8 & 44.7 & 53.0 & 32.0 & 40.8 & 18.4 & 25.2 & 43.8 \\
  \rowcolor{gray!10} \quad \textit{w/} \textbf{\method{}} & \textbf{54.6} & \textbf{42.1} & \textbf{53.8} & \textbf{49.1} & \textbf{45.0} & \textbf{51.1} & \textbf{66.0} & \textbf{57.7} & \textbf{62.0} & \textbf{47.1} & \textbf{59.3} & \textbf{45.7} & \textbf{54.5} & \textbf{60.2} & \textbf{36.4} & \textbf{49.7} & \textbf{28.8} & \textbf{51.5} & \textbf{51.1}\hgain{7.3} \\
  \midrule
  Pretended CoT & 49.0 & 40.0 & 44.6 & 46.7 & 40.3 & 47.2 & 58.6 & 49.5 & 51.9 & 41.9 & 53.0 & 40.9 & 46.0 & 49.3 & 32.2 & 42.1 & 19.6 & 34.0 & 45.3 \\
  \rowcolor{gray!10} \quad \textit{w/} \textbf{\method{}} & \textbf{54.6} & \textbf{42.5} & \textbf{53.3} & \textbf{50.4} & \textbf{45.8} & \textbf{51.2} & \textbf{67.5} & \textbf{59.1} & \textbf{63.4} & \textbf{47.9} & \textbf{59.2} & \textbf{45.7} & \textbf{54.7} & \textbf{56.6} & \textbf{39.2} & \textbf{50.0} & \textbf{28.1} & \textbf{52.7} & \textbf{51.5}\hgain{6.2} \\
  \midrule
  ECHO & 48.4 & 38.4 & 44.3 & 44.2 & 38.4 & 47.5 & 56.5 & 48.3 & 47.1 & 42.4 & 49.0 & 41.9 & 40.5 & 47.5 & 25.8 & 39.9 & 20.7 & 21.5 & 43.7 \\
  \rowcolor{gray!10} \quad \textit{w/} \textbf{\method{}} & \textbf{54.1} & \textbf{40.7} & \textbf{52.3} & \textbf{47.4} & \textbf{42.6} & \textbf{50.7} & \textbf{66.9} & \textbf{57.6} & \textbf{60.6} & \textbf{48.5} & \textbf{55.3} & \textbf{43.9} & \textbf{51.1} & \textbf{55.2} & \textbf{36.0} & \textbf{46.9} & \textbf{23.7} & \textbf{41.2} & \textbf{49.7}\hgain{6.0} \\
  \midrule
  TP & 47.9 & 39.1 & 46.7 & 45.9 & 41.6 & 47.5 & 57.4 & 51.3 & 50.9 & 39.6 & 53.8 & 43.3 & 47.1 & 47.8 & 31.3 & 40.4 & 18.2 & 32.9 & 45.1 \\
  \rowcolor{gray!10} \quad \textit{w/} \textbf{\method{}} & \textbf{54.5} & \textbf{43.1} & \textbf{54.0} & \textbf{50.9} & \textbf{45.7} & \textbf{52.0} & \textbf{67.2} & \textbf{60.2} & \textbf{62.5} & \textbf{47.2} & \textbf{60.6} & \textbf{46.0} & \textbf{56.8} & \textbf{59.2} & \textbf{38.6} & \textbf{48.7} & \textbf{29.3} & \textbf{55.3} & \textbf{51.8}\hgain{6.7} \\
  \midrule
  CP & 47.6 & 39.3 & 44.9 & 45.3 & 39.9 & 46.8 & 57.9 & 50.2 & 50.9 & 41.7 & 53.2 & 42.7 & 46.0 & 51.8 & 32.1 & 41.3 & 18.4 & 36.0 & 44.8 \\
  \rowcolor{gray!10} \quad \textit{w/} \textbf{\method{}} & \textbf{54.9} & \textbf{42.6} & \textbf{54.9} & \textbf{49.1} & \textbf{45.4} & \textbf{51.9} & \textbf{67.2} & \textbf{58.6} & \textbf{62.7} & \textbf{47.6} & \textbf{59.1} & \textbf{45.7} & \textbf{55.1} & \textbf{60.7} & \textbf{37.7} & \textbf{49.3} & \textbf{29.9} & \textbf{53.0} & \textbf{51.6}\hgain{6.8} \\
  \bottomrule
  \end{tabular}}
  \caption{Monolingual performance across 18 languages on MTEB and MMTEB. Each language column reports the average over its monolingual tasks, and AVG denotes the overall mean across all 208 evaluations. Languages marked with $\dagger$ were not used for subspace estimation. The best results are highlighted in \textbf{bold}. \method{} significantly outperforms each corresponding baseline (paired $t$-test, $p<0.01$).}
  \label{tab:mteb}
\end{table*}

\begin{table}[t]
    \centering
%    \scalebox{0.84}{%76
%    \small
 \resizebox{\linewidth}{!}{    
  \tabcolsep=5pt
  \begin{tabular}{lccc|cc}
    \toprule
    \multirow{2}{*}{\textbf{Method}}
        & \multicolumn{3}{c|}{\textbf{FLORES}}
        & \multirow{2}{*}{\textbf{Tatoeba}}
        & \multirow{2}{*}{\textbf{BUCC.v2}} \\
    \cmidrule(lr){2-4}
        & \textbf{En$\rightarrow$X}
        & \textbf{X$\rightarrow$En}
        & \textbf{Mean}
        & & \\
    \midrule
    PromptEOL & 27.42 & 23.43 & 25.43 & 12.95 & 44.25 \\
    \rowcolor{gray!10}
    \quad \textit{w/} \textbf{\method{}}
        & \textbf{48.61} & \textbf{50.16} & \textbf{49.39} & \textbf{47.85} & \textbf{97.13} \\
    \midrule
    Pretended CoT & 30.74 & 24.97 & 27.85 & 16.10 & 49.31 \\
    \rowcolor{gray!10}
    \quad \textit{w/} \textbf{\method{}}
        & \textbf{48.88} & \textbf{50.07} & \textbf{49.47} & \textbf{48.32} & \textbf{97.43} \\
    \midrule
    ECHO & 34.42 & 31.13 & 32.78 & 24.60 & 66.20 \\
    \rowcolor{gray!10}
    \quad \textit{w/} \textbf{\method{}}
        & \textbf{50.34} & \textbf{52.24} & \textbf{51.29} & \textbf{50.49} & \textbf{97.66} \\
    \midrule
    TP & 27.30 & 23.00 & 25.15 & 12.67 & 41.47 \\
    \rowcolor{gray!10}
    \quad \textit{w/} \textbf{\method{}}
        & \textbf{47.12} & \textbf{50.01} & \textbf{48.56} & \textbf{47.51} & \textbf{96.72} \\
    \midrule
    CP & 28.25 & 24.98 & 26.61 & 16.91 & 46.16 \\
    \rowcolor{gray!10}
    \quad \textit{w/} \textbf{\method{}}
        & \textbf{48.54} & \textbf{49.86} & \textbf{49.20} & \textbf{47.82} & \textbf{97.16} \\
    \bottomrule
    \end{tabular}%
    }
    \caption{Cross-lingual bitext mining performance of training-free embedding methods on FLORES, Tatoeba, and BUCC.v2. For FLORES, we report English-to-X and X-to-English results together with their mean. The best results are highlighted in \textbf{bold}.}\vspace{-0.3cm}
    \label{tab:crosslingual}
\end{table}

\subsection{Results}
\paragraph{Monolingual Results.}
Table~\ref{tab:mteb} reports monolingual performance across 18 languages. \method{} consistently improves all evaluated training-free baselines, with average gains of $7.3$, $6.2$, $6.0$, $6.7$, and $6.8$ points for PromptEOL, Pretended CoT, ECHO, TP, and CP, respectively. We hypothesize that these improvements stem from reducing the influence of language identity on embedding similarity, thereby increasing the relative contribution of underlying semantics. Since language-sensitive variation may distort semantic geometry even within a single language, suppressing this effect can make embedding similarity more reflective of semantic relationships, yielding more discriminative representations. Detailed per-dataset results are provided in Appendix~\ref{sec:full-monolingual}.

Importantly, eight evaluation languages, including Japanese (ja), Korean (ko), Bengali (bn), Indonesian (id), Italian (it), Malay (ms), Portuguese (pt), and Telugu (te), are not used for paired-residual subspace estimation. Nevertheless, \method{} consistently enhances embedding performance for these held-out languages. This generalization suggests that the estimated subspaces capture broadly transferable language-sensitive structure rather than patterns specific to the languages used for subspace estimation. 

\paragraph{Cross-Lingual Results.}
Table~\ref{tab:crosslingual} demonstrates that \method{} yields substantially larger improvements in cross-lingual bitext mining than in the monolingual evaluation. Across more than $200$ English-centric language pairs, \method{} improves all five training-free baselines. The gains range from $18.51$ to $23.96$ points on FLORES, $25.89$ to $34.90$ points on Tatoeba, and $31.46$ to $55.25$ points on BUCC.v2. On FLORES, the benefit of \method{} extends to both English-to-X and X-to-English directions, indicating robustness to retrieval direction.

These cross-lingual gains further support our hypothesis that language-specific variation can be disentangled from semantic content to improve embedding quality. By suppressing language-sensitive variation, \method{} produces representations that are less sensitive to language identity and better reflect semantic relationships. Together, these results highlight that \method{} effectively steers frozen LLM representations toward a more language-invariant semantic space, improving multilingual training-free text embeddings.

\section{Ablations and Analyses}
\subsection{Ablations in Subspace Construction}
To examine the importance of \textit{paired-residual} subspace estimation, we compare several alternative construction strategies. \texttt{Baseline} applies no subspace intervention. \texttt{Random} constructs a random orthonormal subspace. Inspired by language-mean subspace estimation for multilingual reasoning~\citep{zhao2025when}, \texttt{Language-Mean} first averages sentence representations within each language and then applies SVD to the centered language-level means. Compared with \method{}, \texttt{Shuffled-Pair} randomly disrupts the original translation pairs before residual construction.
\begin{table}[t]
\centering
\small
\setlength{\tabcolsep}{3pt}
\begin{tabular}{lcccc}
\toprule
\textbf{Method} & \textbf{MTEB} & \textbf{Tatoeba} & \textbf{FLORES} & \textbf{BUCC.v2} \\
\midrule
Baseline       & 46.82 & 12.95 & 25.43 & 44.25 \\
Random         & 46.80 & 12.20 & 24.92 & 43.13 \\
Language-Mean  & 50.00 & 42.95 & 40.90 & 90.91 \\
Shuffled-Pair  & 52.34 & 43.90 & 44.33 & 95.73 \\
\rowcolor{gray!10}
\method{}      & \textbf{54.62} & \textbf{47.85} & \textbf{49.39} & \textbf{97.13} \\
\bottomrule
\end{tabular}
\caption{Performance comparison of different subspace construction strategies using LLaMA-2-7B with PromptEOL. MTEB denotes MTEB (eng, v2). The best results are highlighted in \textbf{bold}.}
\label{tab:components}
\end{table}

As shown in Table~\ref{tab:components}, \texttt{Random} slightly degrades performance compared with \texttt{Baseline}, indicating that the improvements do not simply stem from removing arbitrary low-rank directions. In contrast, \texttt{Language-Mean} yields substantial gains, suggesting that global language-level differences provide useful signals for identifying language-sensitive directions and further validating the effectiveness of language-specific subspace removal.

More importantly, \texttt{Shuffled-Pair} further outperforms \texttt{Language-Mean} despite breaking the correct translation pairs. Since it still relies on residual construction, this comparison highlights the benefit of residual-based subspace estimation beyond language-level averaging. Restoring the correct translation pairs with \method{} leads to the best performance. We attribute this additional gain to the shared semantics within each parallel group, which reduces the effect of unrelated semantic differences when estimating language-sensitive directions. These results confirm the necessity of both residual construction and correct translation pairing in \method{}.

\subsection{Removal versus Amplification}
To further verify the effectiveness of subspace removal, we perform the opposite intervention by amplifying the projected component along the same estimated subspace. Specifically, while \method{} removes $P_\ell=(h_\ell B_\ell)B_\ell^\top$, we instead amplify this component by adding $P_\ell$, $2P_\ell$, or $3P_\ell$ to the last-token hidden states.
\begin{table}[t]
\centering
\footnotesize
\setlength{\tabcolsep}{3pt}
\begin{tabular}{lcccc}
\toprule
\textbf{Intervention} & \textbf{MTEB} & \textbf{Tatoeba} & \textbf{FLORES} & \textbf{BUCC.v2} \\
\midrule
$h_\ell$ (Baseline)         & 46.82 & 12.95 & 25.43 & 44.25 \\
$h_\ell+P_\ell$             & 24.16 & 0.06 & 0.04 & 0.06 \\
$h_\ell+2P_\ell$            & 23.77 & 0.05 & 0.03 & 0.04\\
$h_\ell+3P_\ell$            & 23.64 & 0.05 & 0.03 & 0.03 \\
\rowcolor{gray!10} $h_\ell-P_\ell$ (\method{}) & \textbf{54.62} & \textbf{47.85} & \textbf{49.39} & \textbf{97.13} \\
\bottomrule
\end{tabular}
\caption{Comparison of subspace removal and amplification using LLaMA-2-7B with PromptEOL. MTEB denotes MTEB (eng, v2), and $P_\ell=(h_\ell B_\ell)B_\ell^\top$ is the projection onto the estimated language-sensitive subspace. The best results are highlighted in \textbf{bold}.}
\label{tab:removal-amplification}
\end{table}

As shown in Table~\ref{tab:removal-amplification}, reversing the intervention leads to the opposite effect. Amplifying the estimated subspace significantly degrades performance on Tatoeba, FLORES, and BUCC.v2, whereas removing the same component with \method{} substantially improves all three benchmarks. This sharp contrast indicates that the identified directions encode language-sensitive variation that strongly disrupts cross-lingual semantic alignment when reinforced. The same pattern is also observed on the monolingual English benchmark, where amplification reduces the score from $46.82$ to $24.16$, while removal increases it to $53.81$. These results further confirm the central design of \method{}: suppressing, rather than reinforcing, the estimated language-sensitive subspace leads to more effective semantic representations both across and within languages.

\begin{table}[t]
\centering
\small
\resizebox{\columnwidth}{!}{
\setlength{\tabcolsep}{3pt}
\begin{tabular}{lcccc}
\toprule
\textbf{Method} & \textbf{Mono. Avg.} & \textbf{Tatoeba} & \textbf{FLORES} & \textbf{BUCC.v2} \\
\midrule

\multicolumn{5}{c}{
\texttt{LLaMA-2-7B}
~~\raisebox{-0.2\height}{\includegraphics[height=1em]{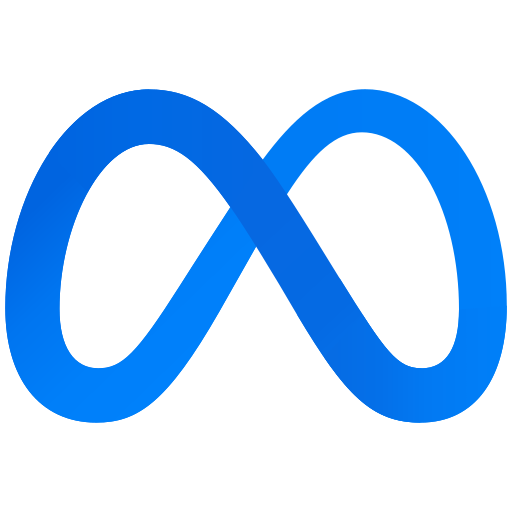}}
}\\
\midrule
PromptEOL
& 43.80 & 12.95 & 25.43 & 44.25 \\
\rowcolor{gray!10}
\quad \textit{w/} \textbf{\method{}}
& \textbf{51.09} & \textbf{47.85} & \textbf{49.39} & \textbf{97.13} \\

\midrule
\multicolumn{5}{c}{
\texttt{Mistral-7B}
~~\raisebox{-0.1\height}{\includegraphics[height=1em]{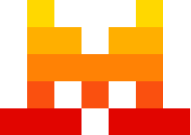}}
}\\
\midrule
PromptEOL
& 49.40 & 29.21 & 37.99 & 49.97 \\
\rowcolor{gray!10}
\quad \textit{w/} \textbf{\method{}}
& \textbf{55.18} & \textbf{56.30} & \textbf{55.96} & \textbf{97.82} \\

\midrule
\multicolumn{5}{c}{
\texttt{Qwen3-8B}
~~\raisebox{-0.2\height}{\includegraphics[height=1em]{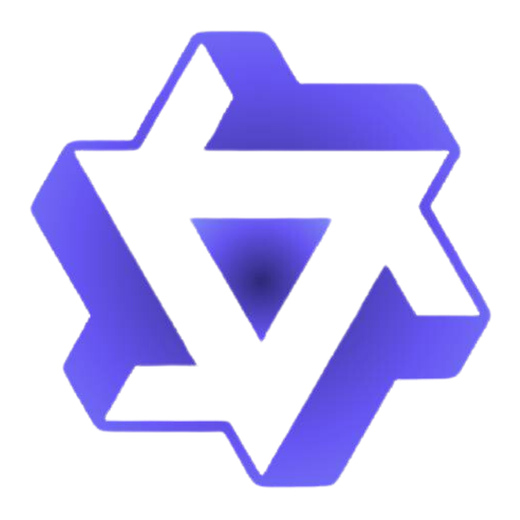}}
}\\
\midrule
PromptEOL
& 51.10 & 32.46 & 33.10 & 47.55 \\
\rowcolor{gray!10}
\quad \textit{w/} \textbf{\method{}}
& \textbf{54.80} & \textbf{59.61} & \textbf{61.78} & \textbf{96.06} \\

\bottomrule
\end{tabular}
}
\caption{Performance of \method{} across different LLM backbones with PromptEOL. Mono. Avg. denotes the average score over 208 monolingual evaluations spanning 18 languages. The best results for each backbone are highlighted in \textbf{bold}.}
\label{tab:backbones}
\end{table}

\subsection{Robustness across Different Backbones}
To examine the generalizability of the proposed paired-residual subspace removal across LLM families, we evaluate \method{} with PromptEOL on LLaMA-2-7B, Mistral-7B, and Qwen3-8B.

As illustrated in Table~\ref{tab:backbones}, \method{} consistently improves PromptEOL across all three LLM backbones on both monolingual and cross-lingual evaluations. These results demonstrate that the effectiveness of \method{} is not specific to a particular LLM family, further suggesting that language-specific variation is broadly present across different decoder-only LLMs.

\subsection{Effect of Subspace Rank}
\label{abl:rank}
By default, we set the subspace rank to $q=K-1$, where $K$ is the number of construction languages. To examine the impact of subspace dimensionality, we vary $q$ from $1$ to $128$ and investigate how many directions are required to effectively capture and remove language-sensitive variation.

\begin{figure}[t]
\centering
\includegraphics[width=\columnwidth]{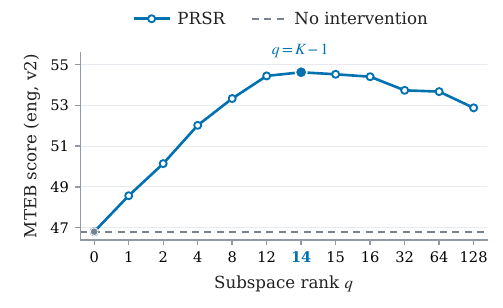}
\caption{Effect of the subspace rank $q$ on MTEB (eng, v2) using LLaMA-2-7B with PromptEOL. Rank 0 denotes the baseline, and $K$ is the number of languages used for paired-residual subspace construction.}
\label{fig:ablation-rank}
\end{figure}

As demonstrated in Figure~\ref{fig:ablation-rank}, performance improves steadily as the rank increases and peaks around $q=14$. This trend suggests that too few directions fail to capture sufficient language-sensitive variation for effective removal. However, the score gradually declines as the rank increases further. Overall, this pattern is consistent with prior findings that language-specific information can be represented in a low-rank subspace~\citep{xie-etal-2022-discovering}. We hypothesize that the performance degradation at higher ranks occurs because the estimated subspace begins to include semantic information useful for embedding extraction. Since the best results are obtained around $q=K-1$, we adopt this setting in all experiments, keeping \method{} simple and avoiding additional hyperparameter tuning.

\subsection{Impact of Intervention Layers}
\label{abl:layer}
To examine the effect of intervention depth, we vary the starting block $\ell_s$ while applying removal at every subsequent block, and evaluate performance on MTEB (eng, v2) across three LLMs families.

\begin{figure}[t]
\centering
\includegraphics[width=\columnwidth]{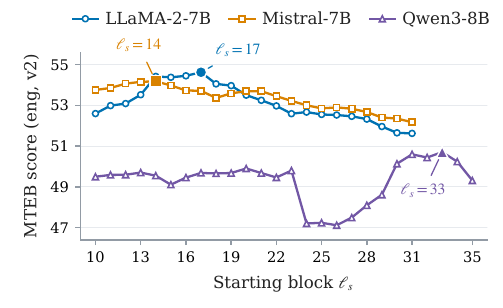}
\caption{Effect of the intervention starting block $\ell_s$ on MTEB (eng, v2) with PromptEOL. Darker markers indicate the best-performing starting block for each LLM.}
\label{fig:ablation-layers}
\end{figure}

As shown in Figure~\ref{fig:ablation-layers}, LLaMA-2-7B and Mistral-7B generally benefit from extending removal from the final layers toward intermediate layers, with the best starting layers at $17$ and $14$, respectively. Extending the intervention further provides no additional benefit. In contrast, Qwen3-8B achieves its highest score when starting at layer $33$, with pronounced fluctuations at earlier starting layers. We suggest that effective steering requires balancing the natural information flow through earlier layers with suppressing language-sensitive variation during later stages of embedding formation. Overall, these results highlight the importance of selecting appropriate intervention ranges for different LLMs.

\section{Conclusion}
In this paper, we presented the first systematic study of multilingual training-free text embeddings in LLMs. Our analysis revealed a language-conditioned embedding geometry in which semantically equivalent texts remain misaligned across languages. To address this issue, we proposed \method{}, an inference-time steering method that estimates language-sensitive subspaces from paired residuals of multilingual parallel texts and suppresses the corresponding components during embedding extraction. Extensive experiments on monolingual and cross-lingual benchmarks demonstrated that \method{} consistently improves existing training-free baselines across different LLM backbones. Further ablations and analyses validated the effectiveness and necessity of the proposed paired-residual subspace removal strategy. We hope this work offers a new perspective on training-free text embeddings beyond English and encourages further exploration into multilingual representation learning and inference-time steering for frozen LLMs.

\section*{Limitations}
Our experiments focus on three decoder-only LLM families ranging from 7B to 8B parameters. Although \method{} is effective across these backbones, its applicability to larger-scale LLMs remains to be further examined. In addition, our cross-lingual evaluation covers English paired with over 200 other languages but does not directly assess semantic matching between non-English languages. Extending the evaluation to these settings would further establish the generalizability of our approach.

\section*{Ethical Considerations}
This work aims to improve multilingual text embeddings from frozen LLMs, with potential benefits for information retrieval and semantic matching across languages. Our experiments use existing pretrained models and benchmark datasets. The resulting representations may retain societal biases from the underlying models and data, and improved semantic alignment does not necessarily ensure fairness across languages or communities. We therefore encourage downstream users to assess potential biases in their intended applications and follow the usage conditions of the models and datasets.

\section*{Acknowledgments}
This study was carried out using the TSUBAME4.0 supercomputer at Institute of Science Tokyo.

\bibliography{main}

@inproceedings{llm2vec,
 title={{LLM}2Vec: Large Language Models Are Secretly Powerful Text Encoders},
 author={Parishad BehnamGhader and Vaibhav Adlakha and Marius Mosbach and Dzmitry Bahdanau and Nicolas Chapados and Siva Reddy},
 booktitle={First Conference on Language Modeling},
 year={2024}
}

@inproceedings{prompteol,
  title={Scaling sentence embeddings with large language models},
  author={Jiang, Ting and Huang, Shaohan and Luan, Zhongzhi and Wang, Deqing and Zhuang, Fuzhen},
  booktitle={Findings of the association for computational linguistics: EMNLP 2024},
  pages={3182--3196},
  year={2024}
}

@inproceedings{echo,
  title={Repetition Improves Language Model Embeddings},
  author={Jacob Mitchell Springer and Suhas Kotha and Daniel Fried and Graham Neubig and Aditi Raghunathan},
  booktitle={The Thirteenth International Conference on Learning Representations},
  year={2025},
  url={https://openreview.net/forum?id=Ahlrf2HGJR}
}

@inproceedings{metaeol,
  title={Meta-task prompting elicits embeddings from large language models},
  author={Lei, Yibin and Wu, Di and Zhou, Tianyi and Shen, Tao and Cao, Yu and Tao, Chongyang and Yates, Andrew},
  booktitle={Proceedings of the 62nd Annual Meeting of the Association for Computational Linguistics (Volume 1: Long Papers)},
  pages={10141--10157},
  year={2024}
}

@inproceedings{kvembedding,
    title = "{KV}-Embedding: Training-free Text Embedding via Internal {KV} Re-routing in Decoder-only {LLM}s",
    author = "Tang, Yixuan  and
      Yang, Yi",
    booktitle = "Proceedings of the 64th Annual Meeting of the {A}ssociation for {C}omputational {L}inguistics (Volume 1: Long Papers)",
    year = "2026",
    pages = "11773--11794"
}

@inproceedings{cp,
  title={Contrastive prompting enhances sentence embeddings in llms through inference-time steering},
  author={Cheng, Zifeng and Wang, Zhonghui and Fu, Yuchen and Jiang, Zhiwei and Yin, Yafeng and Wang, Cong and Gu, Qing},
  booktitle={Proceedings of the 63rd Annual Meeting of the Association for Computational Linguistics (Volume 1: Long Papers)},
  pages={3475--3487},
  year={2025}
}

@inproceedings{tp,
  title={Token prepending: A training-free approach for eliciting better sentence embeddings from llms},
  author={Fu, Yuchen and Cheng, Zifeng and Jiang, Zhiwei and Wang, Zhonghui and Yin, Yafeng and Li, Zhengliang and Gu, Qing},
  booktitle={Proceedings of the 63rd Annual Meeting of the Association for Computational Linguistics (Volume 1: Long Papers)},
  pages={3168--3181},
  year={2025}
}

@article{reverseeol,
  title={ReverseEOL: Improving Training-free Text Embeddings via Text Reversal in Decoder-only LLMs},
  author={Lin, Ailiang and Li, Zhuoyun and Wang, Yusong and Mao, Keyu and Funakoshi, Kotaro and Okumura, Manabu},
  journal={arXiv preprint arXiv:2606.05858},
  year={2026}
}

@inproceedings{tang-etal-2024-language,
    title = "Language-Specific Neurons: The Key to Multilingual Capabilities in Large Language Models",
    author = "Tang, Tianyi  and
      Luo, Wenyang  and
      Huang, Haoyang  and
      Zhang, Dongdong  and
      Wang, Xiaolei  and
      Zhao, Xin  and
      Wei, Furu  and
      Wen, Ji-Rong",
    booktitle = "Proceedings of the 62nd Annual Meeting of the Association for Computational Linguistics (Volume 1: Long Papers)",
    year = "2024",
    pages = "5701--5715"
}

@inproceedings{zhao2024how,
    title={How do Large Language Models Handle Multilingualism?},
    author={Yiran Zhao and Wenxuan Zhang and Guizhen Chen and Kenji Kawaguchi and Lidong Bing},
    booktitle={The Thirty-eighth Annual Conference on Neural Information Processing Systems},
    year={2024},
    url={https://openreview.net/forum?id=ctXYOoAgRy}
}

@inproceedings{maraia-etal-2026-activation,
    title = "Can Activation Steering Generalize Across Languages? A Study on Syllogistic Reasoning in Language Models",
    author = "Maraia, Gabriele  and
      Ranaldi, Leonardo  and
      Valentino, Marco  and
      Zanzotto, Fabio Massimo",
    editor = "Demberg, Vera  and
      Inui, Kentaro  and
      Marquez, Llu{\'i}s",
    booktitle = "Proceedings of the 19th Conference of the {E}uropean Chapter of the {A}ssociation for {C}omputational {L}inguistics (Volume 1: Long Papers)",
    year = "2026",
    pages = "2739--2753"
}

@inproceedings{zhao2025when,
    title={When Less Language is More: Language-Reasoning Disentanglement Makes {LLM}s Better Multilingual Reasoners},
    author={Weixiang Zhao and Jiahe Guo and Yang Deng and Tongtong Wu and Wenxuan Zhang and Yulin Hu and Xingyu Sui and Yanyan Zhao and Wanxiang Che and Bing Qin and Tat-Seng Chua and Ting Liu},
    booktitle={The Thirty-ninth Annual Conference on Neural Information Processing Systems},
    year={2025},
    url={https://openreview.net/forum?id=fleQlZ2VTx}
}

@inproceedings{mteb,
    title = "{MTEB}: Massive Text Embedding Benchmark",
    author = "Muennighoff, Niklas  and
      Tazi, Nouamane  and
      Magne, Loic  and
      Reimers, Nils",
    editor = "Vlachos, Andreas  and
      Augenstein, Isabelle",
    booktitle = "Proceedings of the 17th Conference of the European Chapter of the Association for Computational Linguistics",
    year = "2023",
    pages = "2014--2037"
}

@inproceedings{mmteb,
    title={{MMTEB}: Massive Multilingual Text Embedding Benchmark},
    author={Kenneth Enevoldsen and Isaac Chung and Imene Kerboua and M{\'a}rton Kardos and Ashwin Mathur and David Stap and Jay Gala and Wissam Siblini and Dominik Krzemi{\'n}ski and Genta Indra Winata and Saba Sturua and Saiteja Utpala and Mathieu Ciancone and Marion Schaeffer and Diganta Misra and Shreeya Dhakal and Jonathan Rystr{\o}m and Roman Solomatin and {\"O}mer Veysel {\c{C}}a{\u{g}}atan and Akash Kundu and Martin Bernstorff and Shitao Xiao and Akshita Sukhlecha and Bhavish Pahwa and Rafa{\l} Po{\'s}wiata and Kranthi Kiran GV and Shawon Ashraf and Daniel Auras and Bj{\"o}rn Pl{\"u}ster and Jan Philipp Harries and Lo{\"\i}c Magne and Isabelle Mohr and Dawei Zhu and Hippolyte Gisserot-Boukhlef and Tom Aarsen and Jan Kostkan and Konrad Wojtasik and Taemin Lee and Marek Suppa and Crystina Zhang and Roberta Rocca and Mohammed Hamdy and Andrianos Michail and John Yang and Manuel Faysse and Aleksei Vatolin and Nandan Thakur and Manan Dey and Dipam Vasani and Pranjal A Chitale and Simone Tedeschi and Nguyen Tai and Artem Snegirev and Mariya Hendriksen and Michael G{\"u}nther and Mengzhou Xia and Weijia Shi and Xing Han L{\`u} and Jordan Clive and Gayatri K and Maksimova Anna and Silvan Wehrli and Maria Tikhonova and Henil Shalin Panchal and Aleksandr Abramov and Malte Ostendorff and Zheng Liu and Simon Clematide and Lester James Validad Miranda and Alena Fenogenova and Guangyu Song and Ruqiya Bin Safi and Wen-Ding Li and Alessia Borghini and Federico Cassano and Lasse Hansen and Sara Hooker and Chenghao Xiao and Vaibhav Adlakha and Orion Weller and Siva Reddy and Niklas Muennighoff},
    booktitle={The Thirteenth International Conference on Learning Representations},
    year={2025},
    url={https://openreview.net/forum?id=zl3pfz4VCV}
}

@article{qwen3embedding,
  title={Qwen3 embedding: Advancing text embedding and reranking through foundation models},
  author={Zhang, Yanzhao and Li, Mingxin and Long, Dingkun and Zhang, Xin and Lin, Huan and Yang, Baosong and Xie, Pengjun and Yang, An and Liu, Dayiheng and Lin, Junyang and others},
  journal={arXiv preprint arXiv:2506.05176},
  year={2025}
}

@article{geminiembedding,
  title={Gemini embedding: Generalizable embeddings from gemini},
  author={Lee, Jinhyuk and Chen, Feiyang and Dua, Sahil and Cer, Daniel and Shanbhogue, Madhuri and Naim, Iftekhar and {\'A}brego, Gustavo Hern{\'a}ndez and Li, Zhe and Chen, Kaifeng and Vera, Henrique Schechter and others},
  journal={arXiv preprint arXiv:2503.07891},
  year={2025}
}

@inproceedings{cot,
  title={Simple techniques for enhancing sentence embeddings in generative language models},
  author={Zhang, Bowen and Chang, Kehua and Li, Chunping},
  booktitle={International Conference on Intelligent Computing},
  pages={52--64},
  year={2024},
  organization={Springer}
}

@inproceedings{huang-etal-2025-modular,
    title = "Modular Sentence Encoders: Separating Language Specialization from Cross-Lingual Alignment",
    author = "Huang, Yongxin  and
      Wang, Kexin  and
      Glava{\v{s}}, Goran  and
      Gurevych, Iryna",
    booktitle = "Proceedings of the 63rd Annual Meeting of the Association for Computational Linguistics (Volume 1: Long Papers)",
    year = "2025",
    pages = "2167--2187"
}

@article{artetxe-schwenk-2019-massively,
    title = "Massively Multilingual Sentence Embeddings for Zero-Shot Cross-Lingual Transfer and Beyond",
    author = "Artetxe, Mikel  and
      Schwenk, Holger",
    journal = "Transactions of the Association for Computational Linguistics",
    year = "2019",
    pages = "597--610"
}

@inproceedings{feng-etal-2022-language,
    title = "Language-agnostic {BERT} Sentence Embedding",
    author = "Feng, Fangxiaoyu  and
      Yang, Yinfei  and
      Cer, Daniel  and
      Arivazhagan, Naveen  and
      Wang, Wei",
    booktitle = "Proceedings of the 60th Annual Meeting of the Association for Computational Linguistics (Volume 1: Long Papers)",
    year = "2022",
    pages = "878--891"
}

@inproceedings{yang-etal-2021-simple,
    title = "A Simple and Effective Method To Eliminate the Self Language Bias in Multilingual Representations",
    author = "Yang, Ziyi  and
      Yang, Yinfei  and
      Cer, Daniel  and
      Darve, Eric",
    booktitle = "Proceedings of the 2021 Conference on Empirical Methods in Natural Language Processing",
    year = "2021",
    pages = "5825--5832"
}

@inproceedings{xie-etal-2022-discovering,
    title = "Discovering Low-rank Subspaces for Language-agnostic Multilingual Representations",
    author = "Xie, Zhihui  and
      Zhao, Handong  and
      Yu, Tong  and
      Li, Shuai",
    booktitle = "Proceedings of the 2022 Conference on Empirical Methods in Natural Language Processing",
    year = "2022",
    pages = "5617--5633"
}

@inproceedings{huang-etal-2024-language,
    title = "Language Concept Erasure for Language-invariant Dense Retrieval",
    author = "Huang, Zhiqi  and
      Yu, Puxuan  and
      Ravfogel, Shauli  and
      Allan, James",
    booktitle = "Proceedings of the 2024 Conference on Empirical Methods in Natural Language Processing",
    year = "2024",
    pages = "13261--13273"
}

@inproceedings{dumas-etal-2025-separating,
    title = "Separating Tongue from Thought: Activation Patching Reveals Language-Agnostic Concept Representations in Transformers",
    author = "Dumas, Cl{\'e}ment  and
      Wendler, Chris  and
      Veselovsky, Veniamin  and
      Monea, Giovanni  and
      West, Robert",
    booktitle = "Proceedings of the 63rd Annual Meeting of the Association for Computational Linguistics (Volume 1: Long Papers)",
    year = "2025",
    pages = "31822--31841"
}

@article{llama2,
  title={Llama 2: Open foundation and fine-tuned chat models},
  author={Touvron, Hugo and Martin, Louis and Stone, Kevin and Albert, Peter and Almahairi, Amjad and Babaei, Yasmine and Bashlykov, Nikolay and Batra, Soumya and Bhargava, Prajjwal and Bhosale, Shruti and others},
  journal={arXiv preprint arXiv:2307.09288},
  year={2023}
}

@inproceedings{bucc,
    title = "A Multilingual Dataset for Evaluating Parallel Sentence Extraction from Comparable Corpora",
    author = "Zweigenbaum, Pierre  and
      Sharoff, Serge  and
      Rapp, Reinhard",
    booktitle = "Proceedings of the Eleventh International Conference on Language Resources and Evaluation ({LREC} 2018)",
    year = "2018",
}

@article{flores2,
  title={No language left behind: Scaling human-centered machine translation},
  author={Costa-Juss{\`a}, Marta R and Cross, James and {\c{C}}elebi, Onur and Elbayad, Maha and Heafield, Kenneth and Heffernan, Kevin and Kalbassi, Elahe and Lam, Janice and Licht, Daniel and Maillard, Jean and others},
  journal={arXiv preprint arXiv:2207.04672},
  year={2022}
}

@article{flores1,
    title = "The {F}lores-101 Evaluation Benchmark for Low-Resource and Multilingual Machine Translation",
    author = "Goyal, Naman  and
      Gao, Cynthia  and
      Chaudhary, Vishrav  and
      Chen, Peng-Jen  and
      Wenzek, Guillaume  and
      Ju, Da  and
      Krishnan, Sanjana  and
      Ranzato, Marc{'}Aurelio  and
      Guzm{\'a}n, Francisco  and
      Fan, Angela",
    journal = "Transactions of the Association for Computational Linguistics",
    year = "2022",
    pages = "522--538"
}

@inproceedings{xnli,
    title = "{XNLI}: Evaluating Cross-lingual Sentence Representations",
    author = "Conneau, Alexis  and
      Rinott, Ruty  and
      Lample, Guillaume  and
      Williams, Adina  and
      Bowman, Samuel R.  and
      Schwenk, Holger  and
      Stoyanov, Veselin",
    booktitle = "Proceedings of the 2018 Conference on Empirical Methods in Natural Language Processing",
    year = "2018",
    pages = "2475--2485"
}

@article{mistral,
  title={Mistral 7B},
  author={Jiang, Albert Q and Sablayrolles, Alexandre and Mensch, Arthur and Bamford, Chris and Chaplot, Devendra Singh and Casas, Diego de las and Bressand, Florian and Lengyel, Gianna and Lample, Guillaume and Saulnier, Lucile and others},
  journal={arXiv preprint arXiv:2310.06825},
  year={2023}
}

@article{qwen3,
    title={Qwen3 Technical Report}, 
    author={An Yang and Anfeng Li and Baosong Yang and Beichen Zhang and Binyuan Hui and Bo Zheng and Bowen Yu and Chang Gao and Chengen Huang and Chenxu Lv and Chujie Zheng and Dayiheng Liu and Fan Zhou and Fei Huang and Feng Hu and Hao Ge and Haoran Wei and Huan Lin and Jialong Tang and Jian Yang and Jianhong Tu and Jianwei Zhang and Jianxin Yang and Jiaxi Yang and Jing Zhou and Jingren Zhou and Junyang Lin and Kai Dang and Keqin Bao and Kexin Yang and Le Yu and Lianghao Deng and Mei Li and Mingfeng Xue and Mingze Li and Pei Zhang and Peng Wang and Qin Zhu and Rui Men and Ruize Gao and Shixuan Liu and Shuang Luo and Tianhao Li and Tianyi Tang and Wenbiao Yin and Xingzhang Ren and Xinyu Wang and Xinyu Zhang and Xuancheng Ren and Yang Fan and Yang Su and Yichang Zhang and Yinger Zhang and Yu Wan and Yuqiong Liu and Zekun Wang and Zeyu Cui and Zhenru Zhang and Zhipeng Zhou and Zihan Qiu},
    journal = {arXiv preprint arXiv:2505.09388},
    year={2025}
}

@inproceedings{htp,
    title = "Hierarchical Token Prepending: Enhancing Information Flow in Decoder-based {LLM} Embeddings",
    author = "Ding, Xueying  and
      Huang, Xingyue  and
      Ju, Mingxuan  and
      Collins, Liam  and
      Liu, Yozen  and
      Akoglu, Leman  and
      Shah, Neil  and
      Zhao, Tong",
    booktitle = "Proceedings of the 64th Annual Meeting of the {A}ssociation for {C}omputational {L}inguistics (Volume 1: Long Papers)",
    year = "2026",
    pages = "19047--19066"
}

@inproceedings{lin-etal-2026-embedding,
    title = "Embedding-based In-Context Prompt Training for Enhancing {LLM}s as Text Encoders",
    author = "Lin, Ailiang  and
      Li, Zhuoyun  and
      Mao, Keyu  and
      Funakoshi, Kotaro  and
      Okumura, Manabu",
    booktitle = "Findings of the {A}ssociation for {C}omputational {L}inguistics: {ACL} 2026",
    year = "2026",
    pages = "29079--29095"
}

\clearpage
\appendix

\section{Experimental Details}
\label{sec:appendix-experiments}

\subsection{Monolingual Benchmark Details}
\label{sec:monolingual-benchmarks}
We conduct $208$ monolingual evaluations from MTEB~\citep{mteb} and MMTEB~\citep{mmteb}, spanning $18$ languages: English (en), Chinese (zh), French (fr), Russian (ru), German (de), Japanese (ja), Italian (it), Spanish (es), Portuguese (pt), Arabic (ar), Indonesian (id), Swahili (sw), Vietnamese (vi), Korean (ko), Bengali (bn), Thai (th), Telugu (te), and Malay (ms). The benchmark source and number of datasets for each language are summarized in Table~\ref{tab:monolingual-benchmarks}, with the complete dataset composition provided in Table~\ref{tab:mono-datasets}. For each language, we report the average score over all corresponding evaluations. The overall monolingual average is computed across all $208$ evaluations.

For analysis and ablation studies, we primarily use MTEB (English, v2), as it is one of the most widely adopted benchmarks for evaluating general-purpose text embeddings. It comprises $41$ datasets spanning seven embedding task categories: Retrieval (10), Reranking (2), Clustering (8), Pair Classification (3), Classification (8), Semantic Textual Similarity (9), and Summarization (1).

\subsection{Cross-Lingual Benchmark Details}
\label{sec:crosslingual-benchmarks}
We assess cross-lingual semantic alignment through bitext mining using the \href{https://github.com/embeddings-benchmark/mteb}{MTEB} implementations of BUCC.v2~\citep{bucc}, Tatoeba~\citep{artetxe-schwenk-2019-massively}, and FLORES~\citep{flores1,flores2}. Given a source sentence, the task is to identify its translation among candidate sentences in another language based on embedding similarity. Together, these benchmarks cover English paired with more than $200$ non-English languages. Table~\ref{tab:crosslingual-composition} summarizes their coverage and evaluation settings.

For BUCC.v2 and Tatoeba, we report macro-averaged F1 across $4$ and $112$ X$\rightarrow$En subsets, respectively, where X denotes a non-English language. For FLORES, we evaluate $203$ language--script configurations in both directions and report macro-averaged F1 separately for En$\rightarrow$X and X$\rightarrow$En. The overall FLORES score is the arithmetic mean of the two directional scores.

\begin{table}[t]
\centering
\small
\setlength{\tabcolsep}{7pt}
\begin{tabular}{lllc}
\toprule
\textbf{Language} & \textbf{Code} & \textbf{Benchmark} &
\textbf{\# Datasets} \\
\midrule
English     & en & MTEB (eng, v2) & 41 \\
Chinese     & zh & MTEB (cmn, v1) & 29 \\
French      & fr & MTEB (fra, v1) & 24 \\
Russian     & ru & MTEB (rus, v1) & 18 \\
German      & de & MTEB (deu, v1) & 15 \\
Japanese    & ja & MTEB (jpn, v1) & 13 \\
Korean      & ko & MTEB (kor, v1) & 5  \\
Italian     & it & MMTEB          & 10 \\
Spanish     & es & MMTEB          & 9  \\
Portuguese  & pt & MMTEB          & 7  \\
Arabic      & ar & MMTEB          & 6  \\
Indonesian  & id & MMTEB          & 6  \\
Swahili     & sw & MMTEB          & 5  \\
Vietnamese  & vi & MMTEB          & 5  \\
Bengali     & bn & MMTEB          & 5  \\
Thai        & th & MMTEB          & 4  \\
Telugu      & te & MMTEB          & 4  \\
Malay       & ms & MMTEB          & 2  \\
\midrule
\textbf{Total} & & & \textbf{208} \\
\bottomrule
\end{tabular}
\caption{Benchmark sources and dataset counts for monolingual evaluation.}
\label{tab:monolingual-benchmarks}
\end{table}

\begin{table}[t]
\centering
\small
\setlength{\tabcolsep}{5pt}
\begin{tabular}{lcc}
\toprule
\textbf{Benchmark} & \textbf{\# Languages} & \textbf{Direction} \\
\midrule
BUCC.v2 & 4   & \texttt{X}$\rightarrow$\texttt{En} \\
Tatoeba & 112 & \texttt{X}$\rightarrow$\texttt{En} \\
FLORES  & 195 & \texttt{En}$\rightarrow$\texttt{X} / \texttt{X}$\rightarrow$En \\
\bottomrule
\end{tabular}
\caption{Cross-lingual bitext mining benchmarks. \texttt{En} and \texttt{X} denote English and a non-English language, respectively; \# Languages indicates the number of evaluated non-English languages.}
\label{tab:crosslingual-composition}
\end{table}

\subsection{Implementation Details}
\label{sec:implementation-details}
We conduct experiments with three widely used LLM families: \href{https://huggingface.co/meta-llama/Llama-2-7b-hf}{Llama-2-7b-hf} (LLaMA-2-7B)~\citep{llama2}, \href{https://huggingface.co/mistralai/Mistral-7B-v0.3}{Mistral-7B-v0.3} (Mistral-7B)~\citep{mistral}, and \href{https://huggingface.co/Qwen/Qwen3-8B}{Qwen3-8B}~\citep{qwen3}, with all backbone parameters kept frozen. The maximum input sequence length is set to $512$ tokens for all methods except ECHO, which repeats the input and therefore uses a maximum length of $1024$ tokens. To ensure fair comparison and reproducibility, all experiments use the same evaluation pipeline and random seed.

For subspace estimation, we collect the first $1{,}500$ complete parallel sentence groups across $K=15$ languages from the XNLI validation split~\citep{xnli}, treating premises and hypotheses as separate groups. We estimate layer-specific bases separately for each backbone and training-free baseline with \method{} disabled, following the corresponding input construction and representation extraction procedure. The resulting language-sensitive subspaces remain fixed throughout evaluation and are shared across all evaluation languages.

For input prompting, PromptEOL~\citep{prompteol} uses \textit{``This sentence: [TEXT] means in one word:''}, while Pretended CoT~\citep{cot} extends it with \textit{``After thinking step by step, this sentence: [TEXT] means in one word:''}. For ECHO~\citep{echo}, we follow the improved formulation adopted in ReverseEOL~\citep{reverseeol}, which repeats the input within the PromptEOL-style template: \textit{``Sentence: [TEXT] This sentence: [TEXT] means in one word:''}. In our experiments, CP~\citep{cp} and TP~\citep{tp} adopt PromptEOL as the base prompt while retaining their original best-performing inference settings.

\section{Additional Results}

\subsection{Effect of \method{} on Language-Conditioned Geometry}
\label{sec:prsr-findings-comparison}
To directly assess whether \method{} alleviates the language-conditioned geometry identified in Section~\ref{sec:observation}, we revisit the two motivating findings after steering.

\paragraph{Language clustering.}
We first examine whether the representation space remains strongly organized by language after applying \method{}. We quantify language-specific clustering using normalized mutual information (NMI) and adjusted Rand index (ARI) between unsupervised clusters and language labels. As shown in Table~\ref{tab:prsr-findings-comparison}, \method{} reduces NMI from $0.569$ to $0.148$ and ARI from $0.396$ to $0.108$, showing that the resulting clusters become substantially less aligned with language identity. This suggests that the embedding space becomes less strongly structured by language after steering, consistent with the intended effect of paired-residual subspace removal in suppressing language-specific variation.

\begin{table}[t]
\centering
\small
\setlength{\tabcolsep}{5.3pt}
\begin{tabular}{lcc}
\toprule
\textbf{Measure} & \textbf{Baseline} & \textit{w/} \textbf{\method{}} \\
\midrule
NMI $\downarrow$ & 0.569 & \textbf{0.148} \\
ARI $\downarrow$ & 0.396 & \textbf{0.108} \\
\midrule
Ordering violations (\%) $\downarrow$ & 36.24 & \textbf{10.06} \\
Distance ratio (\%) $\downarrow$ & 85.56 & \textbf{50.98} \\
Translation distance & 0.203 & 0.254 \\
Same-language negative distance & 0.237 & 0.499 \\
\bottomrule
\end{tabular}
\caption{Effect of \method{} on the two motivating findings in Section~\ref{sec:observation} using LLaMA-2-7B with PromptEOL.}
\label{tab:prsr-findings-comparison}
\end{table}

\paragraph{Semantic correspondence.}
We next examine whether semantic equivalence across languages is more faithfully reflected in embedding similarity after applying \method{}. Following the analysis in Section~\ref{sec:observation}, we compare each anchor with its translation and same-language negatives. As shown in Table~\ref{tab:prsr-findings-comparison}, the proportion of cases where a same-language negative is more similar than the corresponding translation decreases substantially from $36.24\%$ to $10.06\%$. Meanwhile, the translation-to-negative distance ratio decreases from $85.56\%$ to $50.98\%$, indicating a much clearer separation between semantically equivalent translations and semantically different sentences. Notably, this improvement does not result from simply pulling translations closer together. Instead, the distance to same-language negatives increases substantially more than the distance to translations, making embedding similarity more reflective of semantic correspondence than shared language identity.

\begin{figure}[t]
\centering
\includegraphics[width=\columnwidth]{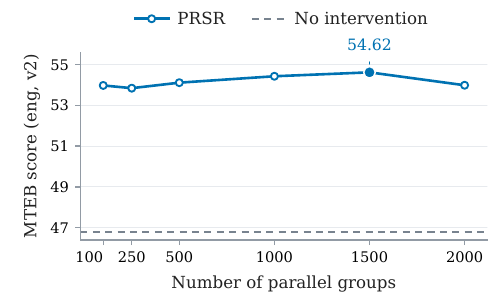}
\caption{Effect of the number of parallel sentence groups used for subspace estimation on MTEB (eng, v2) using LLaMA-2-7B with PromptEOL.}
\label{fig:ablation-sample-size}
\end{figure}
\subsection{Effect of Construction Sample Size}
To examine the data requirements of \method{}, we vary the number of parallel sentence groups sampled from the XNLI validation set~\citep{xnli} for paired-residual subspace estimation from $100$ to $2{,}000$ and evaluate the resulting embeddings on MTEB (eng, v2) using LLaMA-2-7B with PromptEOL.

As illustrated in Figure~\ref{fig:ablation-sample-size}, using only $100$ groups already yields a substantial improvement over the baseline, with performance comparable to that obtained using $2{,}000$ groups. The highest score is achieved with $1{,}500$ groups, while further increasing the construction sample size to $2{,}000$ provides no additional benefit. More importantly, all six settings achieve similar performance and significantly outperform the baseline, suggesting that \method{} is not highly sensitive to the amount of construction data. These results further indicate that effective language-sensitive subspaces can be estimated from relatively few parallel groups, without requiring a large construction corpus.

\begin{figure}[t]
\centering
\includegraphics[width=\columnwidth]{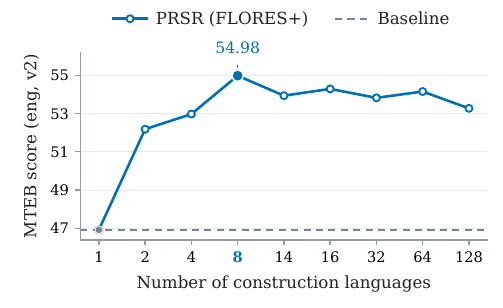}
\caption{Effect of the subspace construction corpus and the number of construction languages on MTEB (eng, v2) using LLaMA-2-7B with PromptEOL.}
\label{fig:construction-data}
\end{figure}
\subsection{Effect of Subspace Construction Data}
In this section, we further investigate how the choice of subspace construction data affects \method{}, considering both the source of parallel sentences and the number of languages used for subspace estimation. All experiments are evaluated on MTEB (eng, v2) using LLaMA-2-7B with PromptEOL.

\paragraph{Construction corpus.}
We first examine whether \method{} depends strongly on the choice of parallel corpus by replacing XNLI with FLORES+ for subspace construction. As shown in Figure~\ref{fig:construction-data}, \method{} achieves comparable performance with the two data sources, obtaining $54.62$ with XNLI and $54.98$ with FLORES+. This result suggests that the effectiveness of \method{} is not specific to a particular parallel corpus and that effective language-specific subspaces can be estimated from different sources of multilingual parallel data.

\paragraph{Number of construction languages.}
We next investigate the effect of the number of languages used for subspace estimation. With its broad multilingual coverage, FLORES+ provides a suitable setting for systematically varying the number of construction languages. As shown in Figure~\ref{fig:construction-data}, using only two languages already significantly improves over the unsteered baseline, increasing the MTEB score from $46.93$ to $52.18$. Performance improves as more languages are introduced and reaches its highest value of $54.98$ with eight languages. Further increasing the number of construction languages does not yield additional gains, with performance remaining around $53$--$54$ across most larger settings. These results suggest that incorporating multilingual diversity is important for estimating effective language-sensitive subspaces, while further increasing the number of languages provides limited benefit once sufficient cross-lingual variation has been captured. For simplicity and to avoid introducing an additional language-selection hyperparameter, we use all $15$ languages available in XNLI for subspace construction throughout our experiments.

\begin{table}[t]
\centering
\small
\setlength{\tabcolsep}{2.5pt}
\begin{tabular}{lcccc}
\toprule
\textbf{Method} & \textbf{Mono. Avg.} & \textbf{Tatoeba} & \textbf{FLORES} & \textbf{BUCC.v2} \\
\midrule
ECHO
& 41.51 & 1.44 & 1.63 & 0.50 \\
\rowcolor{gray!10}
 \quad \textit{w/} \textbf{\method{}}
& \textbf{44.51} & \textbf{20.93} & \textbf{19.56} & \textbf{91.79} \\
\bottomrule
\end{tabular}
\caption{Comparison of ECHO with and without \method{} under mean pooling using LLaMA-2-7B. Mono. Avg. denotes the average score across all $208$ monolingual evaluations.}
\label{tab:mean-pooling}
\end{table}
\subsection{Compatibility with Mean Pooling}
\method{} is primarily formulated for training-free baselines that extract embeddings from the last-token representation, with both paired-residual subspace construction and inference-time steering operating on last-token hidden states. In this section, we further examine whether \method{} remains effective under another commonly used embedding extraction strategy: mean pooling. 

Here we use the original ECHO formulation~\citep{echo}, which repeats the input and obtains the text embedding by mean pooling the tokens of the second occurrence. Specifically, unlike the PromptEOL-style ECHO variant used in our main experiments, we use the official prompt \textit{``Rewrite the following paragraph: [TEXT]. The rewritten paragraph: [TEXT]''}. To construct the language-sensitive subspace at each selected layer, we first average the tokens of the second occurrence into a sentence-level vector and then construct paired residuals across parallel sentences following the same procedure as in Section~\ref{main:setup}. During inference, we apply the same projection removal to every token of the second occurrence at each selected layer.

As shown in Table~\ref{tab:mean-pooling}, \method{} substantially improves ECHO under mean pooling. The average monolingual score across the $208$ evaluations increases from $41.5$ to $44.5$. The improvements are considerably larger on cross-lingual bitext mining, with Tatoeba increasing from $1.44$ to $20.93$, FLORES from $1.63$ to $19.56$, and BUCC from $0.50$ to $91.79$. These results provide further evidence that language-specific variation is broadly present in training-free text embeddings and is not specific to last-token pooling. More importantly, the strong performance under mean pooling shows that \method{} can be effectively extended to different pooling strategies through corresponding adaptations to subspace construction and inference-time steering, demonstrating the broader applicability of the proposed paired-residual subspace removal strategy.

\subsection{Full Monolingual Results}
\label{sec:full-monolingual}
Tables~\ref{tab:full-english-results}--\ref{tab:full-nonenglish-results-3} report the complete per-dataset results for all $208$ monolingual evaluations, comparing each training-free baseline with and without \method{}. The $41$ English datasets from MTEB (eng, v2) are grouped into seven task categories, while the remaining $167$ non-English evaluations are grouped by language.

\begin{table*}[t]
\centering
\scriptsize
\setlength{\tabcolsep}{4pt}
\begin{tabular}{p{0.14\textwidth} c p{0.75\textwidth}}
\toprule
\textbf{Language} & \textbf{\#} & \textbf{Datasets} \\
\midrule

\textbf{English (en)} & 41 &
ArguAna, ArXivHierarchicalClusteringP2P, ArXivHierarchicalClusteringS2S,
AskUbuntuDupQuestions, Banking77Classification, BIOSSES,
BiorxivClusteringP2P.v2, ClimateFEVERHardNegatives,
CQADupstackGamingRetrieval, CQADupstackUnixRetrieval,
FEVERHardNegatives, FiQA2018, HotpotQAHardNegatives,
ImdbClassification, MassiveIntentClassification,
MassiveScenarioClassification, MedrxivClusteringP2P.v2,
MedrxivClusteringS2S.v2, MTOPDomainClassification,
MindSmallReranking, SCIDOCS, SICK-R, SprintDuplicateQuestions,
StackExchangeClusteringP2P.v2, StackExchangeClustering.v2,
STS12, STS13, STS14, STS15, STS17, STS22.v2, STSBenchmark,
TRECCOVID, Touche2020Retrieval.v3, ToxicConversationsClassification,
TweetSentimentExtractionClassification, TwentyNewsgroupsClustering.v2,
TwitterSemEval2015, TwitterURLCorpus,
AmazonCounterfactualClassification, SummEvalSummarization.v2 \\
\midrule

\textbf{Chinese (zh)} & 29 &
CMedQAv1-reranking, CMedQAv2-reranking, MMarcoReranking,
T2Reranking, CLSClusteringS2S, CLSClusteringP2P,
ThuNewsClusteringS2S, ThuNewsClusteringP2P, LCQMC, PAWSX,
QBQTC, Waimai, OnlineShopping, JDReview, MultilingualSentiment,
ATEC, BQ, STSB, AFQMC, Cmnli, Ocnli, CmedqaRetrieval,
CovidRetrieval, DuRetrieval, EcomRetrieval, MMarcoRetrieval,
MedicalRetrieval, T2Retrieval, VideoRetrieval \\
\midrule

\textbf{French (fr)} & 24 &
AmazonReviewsClassification, MasakhaNEWSClassification,
MassiveIntentClassification, MassiveScenarioClassification,
MTOPDomainClassification, MTOPIntentClassification,
AlloProfClusteringS2S, AlloProfClusteringP2P, HALClusteringS2S,
MasakhaNEWSClusteringP2P, MasakhaNEWSClusteringS2S,
MLSUMClusteringP2P, MLSUMClusteringS2S, AlloprofReranking,
SyntecReranking, AlloprofRetrieval, BSARDRetrieval,
MintakaRetrieval, SyntecRetrieval, XPQARetrieval, SICKFr,
STSBenchmarkMultilingualSTS, STS22, SummEvalFr \\
\midrule

\textbf{Russian (ru)} & 18 &
GeoreviewClassification, HeadlineClassification,
InappropriatenessClassification, KinopoiskClassification,
MassiveIntentClassification, MassiveScenarioClassification,
RuReviewsClassification, RuSciBenchGRNTIClassification,
RuSciBenchOECDClassification, GeoreviewClusteringP2P,
RuSciBenchGRNTIClusteringP2P, RuSciBenchOECDClusteringP2P,
RuBQReranking, MIRACLReranking, RuBQRetrieval,
RUParaPhraserSTS, STS22, RuSTSBenchmarkSTS \\
\midrule

\textbf{German (de)} & 15 &
AmazonCounterfactualClassification, AmazonReviewsClassification,
MTOPDomainClassification, MTOPIntentClassification,
MassiveIntentClassification, MassiveScenarioClassification,
BlurbsClusteringP2P, BlurbsClusteringS2S,
TenKGnadClusteringP2P, TenKGnadClusteringS2S,
GermanQuAD-Retrieval, XMarket, GerDaLIR,
GermanSTSBenchmark, STS22 \\
\midrule

\textbf{Japanese (ja)} & 13 &
LivedoorNewsClustering.v2, MewsC16JaClustering,
AmazonReviewsClassification, AmazonCounterfactualClassification,
MassiveIntentClassification, MassiveScenarioClassification,
JSICK, JaqketRetrieval, JaGovFaqsRetrieval,
NLPJournalAbsIntroRetrieval, NLPJournalTitleAbsRetrieval,
NLPJournalTitleIntroRetrieval, ESCIReranking \\
\midrule

\textbf{Italian (it)} & 10 &
ItaCaseholdClassification, MassiveIntentClassification,
MultiHateClassification, SwissJudgementClassification,
SIB200ClusteringS2S, BelebeleRetrieval,
WikipediaRetrievalMultilingual, RTE3,
WikipediaRerankingMultilingual, STS22.v2 \\
\midrule

\textbf{Spanish (es)} & 9 &
CataloniaTweetClassification, MassiveIntentClassification,
MultiHateClassification, SIB200ClusteringS2S,
BelebeleRetrieval, MLQARetrieval, XNLI, STS17, STS22.v2 \\
\midrule

\textbf{Portuguese (pt)} & 7 &
AfriSentiClassification, MassiveIntentClassification,
MultiHateClassification, SIB200ClusteringS2S,
BelebeleRetrieval, WikipediaRetrievalMultilingual,
WikipediaRerankingMultilingual \\
\midrule

\textbf{Arabic (ar)} & 6 &
MassiveIntentClassification, MultiHateClassification,
MLQARetrieval, XNLI, STS17, STS22.v2 \\
\midrule

\textbf{Indonesian (id)} & 6 &
MassiveIntentClassification, NusaX-senti,
SIB200ClusteringS2S, indonli, BelebeleRetrieval, SemRel24STS \\
\midrule

\textbf{Swahili (sw)} & 5 &
AfriSentiClassification, MasakhaNEWSClassification,
MassiveIntentClassification, MasakhaNEWSClusteringS2S, XNLI \\
\midrule

\textbf{Vietnamese (vi)} & 5 &
MassiveIntentClassification, SIB200ClusteringS2S,
BelebeleRetrieval, MLQARetrieval, XNLI \\
\midrule

\textbf{Korean (ko)} & 5 &
KorSTS, KLUE-STS, KLUE-TC, MIRACLReranking, Ko-StrategyQA \\
\midrule

\textbf{Bengali (bn)} & 5 &
MassiveIntentClassification, SIB200ClusteringS2S,
WikipediaRerankingMultilingual, BelebeleRetrieval,
WikipediaRetrievalMultilingual \\
\midrule

\textbf{Thai (th)} & 4 &
MassiveIntentClassification, SIB200ClusteringS2S,
XNLI, BelebeleRetrieval \\
\midrule

\textbf{Telugu (te)} & 4 &
MassiveIntentClassification, SIB200ClusteringS2S,
BelebeleRetrieval, SemRel24STS \\
\midrule

\textbf{Malay (ms)} & 2 &
SIB200ClusteringS2S, BelebeleRetrieval \\
\midrule

\textbf{Overall} & \textbf{208} & \\

\bottomrule
\end{tabular}
\caption{Dataset composition of the monolingual evaluation across 18 languages.}
\label{tab:mono-datasets}
\end{table*}

\begin{table*}[t]
\centering
\scriptsize
\setlength{\tabcolsep}{2.5pt}
\renewcommand{\arraystretch}{0.90}
\resizebox{\textwidth}{!}{%
\begin{tabular}{lcccccccccc}
\toprule
\textbf{Dataset}
& \multicolumn{2}{c}{\textbf{PromptEOL}}
& \multicolumn{2}{c}{\textbf{Pretended CoT}}
& \multicolumn{2}{c}{\textbf{ECHO}}
& \multicolumn{2}{c}{\textbf{CP}}
& \multicolumn{2}{c}{\textbf{TP}} \\
\cmidrule(lr){2-3}
\cmidrule(lr){4-5}
\cmidrule(lr){6-7}
\cmidrule(lr){8-9}
\cmidrule(lr){10-11}
& Base & \textit{w/} \textbf{\method{}}
& Base & \textit{w/} \textbf{\method{}}
& Base & \textit{w/} \textbf{\method{}}
& Base & \textit{w/} \textbf{\method{}}
& Base & \textit{w/} \textbf{\method{}} \\
\midrule

\multicolumn{11}{c}{\textit{Retrieval (10)}} \\
\midrule
ArguAna & 27.08 & 41.25 & 30.51 & 44.67 & 26.18 & 42.61 & 26.92 & 42.60 & 22.22 & 39.81 \\
ClimateFEVERHardNegatives & 5.40 & 13.10 & 4.88 & 14.02 & 7.95 & 14.93 & 5.70 & 13.79 & 4.36 & 11.30 \\
CQADupstackGamingRetrieval & 13.92 & 32.20 & 18.73 & 32.06 & 17.55 & 31.15 & 16.42 & 33.28 & 17.66 & 35.29 \\
CQADupstackUnixRetrieval & 16.77 & 27.55 & 17.53 & 26.63 & 16.32 & 25.92 & 17.43 & 27.91 & 16.77 & 26.72 \\
FEVERHardNegatives & 7.46 & 29.17 & 8.41 & 22.93 & 17.23 & 41.96 & 7.88 & 28.20 & 6.89 & 24.86 \\
FiQA2018 & 9.77 & 21.65 & 13.77 & 23.13 & 10.24 & 21.10 & 10.75 & 21.33 & 8.58 & 19.79 \\
HotpotQAHardNegatives & 6.06 & 22.40 & 9.62 & 23.12 & 15.14 & 27.80 & 8.13 & 22.78 & 7.38 & 24.07 \\
SCIDOCS & 9.82 & 14.83 & 10.35 & 13.60 & 4.93 & 11.63 & 11.13 & 15.42 & 9.28 & 13.56 \\
Touche2020Retrieval.v3 & 7.04 & 26.21 & 15.60 & 28.48 & 11.43 & 27.00 & 9.98 & 25.37 & 12.38 & 27.21 \\
TRECCOVID & 23.08 & 42.75 & 35.88 & 46.43 & 37.21 & 44.33 & 26.03 & 44.67 & 33.48 & 49.40 \\
\cmidrule(lr){1-11}
\textbf{Avg.} & 12.64 & \textbf{27.11} & 16.53 & \textbf{27.51} & 16.42 & \textbf{28.84} & 14.04 & \textbf{27.53} & 13.90 & \textbf{27.20} \\
\midrule

\multicolumn{11}{c}{\textit{Reranking (2)}} \\
\midrule
AskUbuntuDupQuestions & 53.13 & 59.45 & 56.77 & 59.43 & 56.57 & 59.03 & 54.02 & 59.64 & 55.56 & 59.22 \\
MindSmallReranking & 29.00 & 30.95 & 29.99 & 30.60 & 28.81 & 30.86 & 29.81 & 30.76 & 29.88 & 31.01 \\
\cmidrule(lr){1-11}
\textbf{Avg.} & 41.06 & \textbf{45.20} & 43.38 & \textbf{45.02} & 42.69 & \textbf{44.94} & 41.91 & \textbf{45.20} & 42.72 & \textbf{45.12} \\
\midrule

\multicolumn{11}{c}{\textit{Clustering (8)}} \\
\midrule
ArXivHierarchicalClusteringP2P & 54.88 & 57.66 & 56.73 & 58.46 & 56.68 & 58.93 & 55.19 & 55.87 & 56.56 & 58.40 \\
ArXivHierarchicalClusteringS2S & 55.44 & 58.79 & 55.83 & 55.86 & 55.09 & 57.78 & 55.77 & 57.62 & 57.05 & 56.71 \\
BiorxivClusteringP2P.v2 & 21.95 & 28.65 & 24.21 & 28.13 & 25.23 & 29.16 & 21.73 & 29.62 & 22.60 & 28.14 \\
MedrxivClusteringP2P.v2 & 25.88 & 28.00 & 25.55 & 27.80 & 25.98 & 26.77 & 25.38 & 28.76 & 25.15 & 27.37 \\
MedrxivClusteringS2S.v2 & 28.44 & 29.84 & 28.67 & 28.95 & 26.65 & 29.16 & 28.75 & 30.59 & 27.78 & 29.78 \\
StackExchangeClustering.v2 & 51.90 & 54.95 & 54.69 & 55.50 & 51.02 & 55.35 & 52.79 & 54.89 & 53.70 & 56.08 \\
StackExchangeClusteringP2P.v2 & 33.87 & 35.71 & 34.42 & 35.81 & 34.20 & 35.57 & 34.16 & 36.15 & 33.92 & 34.70 \\
TwentyNewsgroupsClustering.v2 & 29.25 & 47.71 & 34.14 & 45.46 & 28.88 & 45.43 & 31.29 & 48.24 & 33.17 & 47.54 \\
\cmidrule(lr){1-11}
\textbf{Avg.} & 37.70 & \textbf{42.66} & 39.28 & \textbf{42.00} & 37.97 & \textbf{42.27} & 38.13 & \textbf{42.72} & 38.74 & \textbf{42.34} \\
\midrule

\multicolumn{11}{c}{\textit{Pair Classification (3)}} \\
\midrule
SprintDuplicateQuestions & 38.28 & 49.07 & 41.06 & 47.57 & 42.77 & 45.62 & 39.54 & 50.47 & 45.01 & 50.47 \\
TwitterSemEval2015 & 65.64 & 71.97 & 68.41 & 72.75 & 65.98 & 68.24 & 67.00 & 72.59 & 65.46 & 71.62 \\
TwitterURLCorpus & 76.22 & 84.87 & 79.68 & 85.01 & 81.27 & 85.21 & 77.04 & 84.86 & 75.29 & 84.26 \\
\cmidrule(lr){1-11}
\textbf{Avg.} & 60.05 & \textbf{68.64} & 63.05 & \textbf{68.44} & 63.34 & \textbf{66.36} & 61.19 & \textbf{69.30} & 61.92 & \textbf{68.78} \\
\midrule

\multicolumn{11}{c}{\textit{Classification (8)}} \\
\midrule
AmazonCounterfactualClassification & 69.46 & 70.25 & 70.52 & 70.81 & 73.39 & 73.70 & 69.07 & 69.28 & 66.79 & 67.73 \\
Banking77Classification & 78.29 & 83.10 & 79.13 & 83.22 & 79.43 & 82.46 & 78.87 & 83.24 & 79.37 & 83.37 \\
ImdbClassification & 92.25 & 92.97 & 90.43 & 91.53 & 91.36 & 90.49 & 92.30 & 93.16 & 92.05 & 93.14 \\
MassiveIntentClassification & 73.09 & 75.66 & 73.70 & 75.61 & 74.27 & 74.99 & 73.59 & 75.95 & 73.82 & 75.48 \\
MassiveScenarioClassification & 75.77 & 78.84 & 76.10 & 77.82 & 76.90 & 78.20 & 76.35 & 78.88 & 77.09 & 78.45 \\
MTOPDomainClassification & 90.21 & 93.36 & 90.74 & 93.22 & 90.73 & 92.67 & 90.46 & 93.46 & 90.99 & 93.67 \\
ToxicConversationsClassification & 65.13 & 67.62 & 64.40 & 65.65 & 65.55 & 64.35 & 65.69 & 67.26 & 66.40 & 67.40 \\
TweetSentimentExtractionClassification & 59.77 & 59.99 & 60.58 & 59.60 & 61.71 & 59.41 & 59.79 & 59.44 & 59.42 & 59.33 \\
\cmidrule(lr){1-11}
\textbf{Avg.} & 75.50 & \textbf{77.72} & 75.70 & \textbf{77.18} & 76.67 & \textbf{77.03} & 75.77 & \textbf{77.59} & 75.74 & \textbf{77.32} \\
\midrule

\multicolumn{11}{c}{\textit{STS (9)}} \\
\midrule
BIOSSES & 63.02 & 75.23 & 74.04 & 78.00 & 62.76 & 73.36 & 65.43 & 77.43 & 64.48 & 76.42 \\
SICK-R & 70.14 & 75.42 & 70.48 & 75.51 & 70.72 & 74.16 & 70.91 & 75.92 & 72.40 & 75.60 \\
STS12 & 58.86 & 68.23 & 61.11 & 68.95 & 60.32 & 64.19 & 60.84 & 68.48 & 60.60 & 67.49 \\
STS13 & 76.96 & 83.40 & 79.63 & 84.58 & 76.56 & 81.12 & 78.56 & 83.77 & 78.35 & 83.72 \\
STS14 & 66.35 & 74.59 & 69.41 & 76.42 & 65.02 & 70.58 & 68.46 & 75.34 & 68.65 & 75.18 \\
STS15 & 73.26 & 81.65 & 74.79 & 82.48 & 75.67 & 80.09 & 73.90 & 82.35 & 75.20 & 82.11 \\
STS17 & 76.10 & 85.55 & 74.98 & 85.71 & 77.19 & 82.68 & 76.47 & 85.93 & 77.78 & 86.31 \\
STS22.v2 & 40.73 & 49.91 & 43.57 & 51.60 & 43.99 & 53.70 & 40.34 & 51.04 & 39.42 & 46.46 \\
STSBenchmark & 71.70 & 80.59 & 71.98 & 80.98 & 70.40 & 77.85 & 72.62 & 80.67 & 72.96 & 80.63 \\
\cmidrule(lr){1-11}
\textbf{Avg.} & 66.35 & \textbf{74.95} & 68.89 & \textbf{76.03} & 66.96 & \textbf{73.08} & 67.50 & \textbf{75.66} & 67.76 & \textbf{74.88} \\
\midrule

\multicolumn{11}{c}{\textit{Summarization (1)}} \\
\midrule
SummEvalSummarization.v2 & 28.25 & \textbf{34.15} & 28.50 & \textbf{30.92} & 24.82 & \textbf{28.33} & 26.87 & \textbf{32.08} & 27.21 & \textbf{32.83} \\
\midrule
\textbf{Overall (41)} & 46.80 & \textbf{54.62} & 49.01 & \textbf{54.61} & 48.39 & \textbf{54.09} & 47.64 & \textbf{54.86} & 47.88 & \textbf{54.45} \\
\bottomrule
\end{tabular}%
}
\caption{Detailed results of training-free baselines with and without \method{} on MTEB (eng, v2) using LLaMA-2-7B. For category-level and overall averages, the better result between each baseline and its \method{}-enhanced version is highlighted in \textbf{bold}.}
\label{tab:full-english-results}
\end{table*}

\begin{table*}[t]
\centering
\scriptsize
\setlength{\tabcolsep}{3.2pt}
\begin{tabular}{llcccccccccc}
\toprule
\textbf{Lang.} & \textbf{Dataset}
& \multicolumn{2}{c}{\textbf{PromptEOL}}
& \multicolumn{2}{c}{\textbf{Pretended CoT}}
& \multicolumn{2}{c}{\textbf{ECHO}}
& \multicolumn{2}{c}{\textbf{CP}}
& \multicolumn{2}{c}{\textbf{TP}} \\
\cmidrule(lr){3-4}
\cmidrule(lr){5-6}
\cmidrule(lr){7-8}
\cmidrule(lr){9-10}
\cmidrule(lr){11-12}
& & Base & \textit{w/} \textbf{\method{}}
& Base & \textit{w/} \textbf{\method{}}
& Base & \textit{w/} \textbf{\method{}}
& Base & \textit{w/} \textbf{\method{}}
& Base & \textit{w/} \textbf{\method{}} \\
\midrule

% ==================== Japanese ====================
\multirow{13}{*}{ja}
& LivedoorNewsClustering.v2 & 25.76 & 32.58 & 26.70 & 32.99 & 19.73 & 31.36 & 28.49 & 34.74 & 25.36 & 34.09 \\
& MewsC16JaClustering & 45.46 & 48.54 & 46.25 & 47.28 & 41.21 & 43.67 & 44.62 & 50.19 & 43.93 & 46.73 \\
& AmazonReviewsClassification & 38.07 & 38.06 & 39.53 & 38.84 & 39.73 & 38.22 & 37.75 & 37.61 & 37.65 & 36.96 \\
& AmazonCounterfactualClassification & 70.63 & 70.37 & 70.66 & 70.94 & 72.28 & 72.77 & 70.36 & 70.12 & 69.95 & 70.40 \\
& MassiveIntentClassification & 70.08 & 71.63 & 70.26 & 71.44 & 70.58 & 71.81 & 70.37 & 71.96 & 70.53 & 71.76 \\
& MassiveScenarioClassification & 73.87 & 75.36 & 73.99 & 75.31 & 74.44 & 75.99 & 74.44 & 75.79 & 75.25 & 75.48 \\
& JSICK & 66.40 & 73.43 & 68.12 & 75.91 & 67.27 & 72.89 & 68.74 & 75.27 & 71.99 & 76.31 \\
& JaqketRetrieval & 27.12 & 34.77 & 27.98 & 34.48 & 29.64 & 30.01 & 25.26 & 34.70 & 33.87 & 35.83 \\
& JaGovFaqsRetrieval & 37.41 & 48.51 & 41.57 & 49.21 & 40.48 & 49.23 & 39.43 & 48.81 & 39.12 & 49.39 \\
& NLPJournalAbsIntroRetrieval & 30.03 & 29.77 & 28.48 & 28.29 & 26.06 & 29.88 & 29.00 & 31.08 & 30.57 & 31.06 \\
& NLPJournalTitleAbsRetrieval & 29.10 & 43.78 & 27.09 & 41.80 & 38.87 & 43.67 & 28.14 & 46.47 & 27.21 & 49.40 \\
& NLPJournalTitleIntroRetrieval & 11.58 & 16.05 & 13.16 & 17.19 & 16.54 & 17.94 & 12.08 & 16.95 & 12.38 & 16.93 \\
& ESCIReranking & 79.70 & 81.34 & 80.11 & 81.88 & 80.26 & 81.47 & 79.86 & 81.39 & 80.09 & 81.50 \\
\cmidrule(lr){2-12}
& \textbf{Avg.} & 46.60 & \textbf{51.09} & 47.22 & \textbf{51.20} & 47.47 & \textbf{50.68} & 46.81 & \textbf{51.93} & 47.53 & \textbf{51.99} \\
\midrule

% ==================== German ====================
\multirow{15}{*}{de}
& AmazonCounterfactualClassification & 67.06 & 66.81 & 67.73 & 68.04 & 67.72 & 67.13 & 67.30 & 66.22 & 67.42 & 66.58 \\
& AmazonReviewsClassification & 41.43 & 41.64 & 42.42 & 42.14 & 41.58 & 40.57 & 41.00 & 41.33 & 42.03 & 41.99 \\
& MTOPDomainClassification & 86.33 & 90.10 & 87.46 & 90.34 & 87.45 & 88.69 & 87.22 & 90.71 & 88.88 & 90.95 \\
& MTOPIntentClassification & 75.54 & 78.90 & 76.53 & 79.25 & 76.88 & 78.42 & 76.04 & 79.10 & 77.15 & 79.34 \\
& MassiveIntentClassification & 63.54 & 66.72 & 65.34 & 67.34 & 65.61 & 65.79 & 64.61 & 67.67 & 65.86 & 67.47 \\
& MassiveScenarioClassification & 68.62 & 71.93 & 69.26 & 71.93 & 69.82 & 70.24 & 69.68 & 72.95 & 70.81 & 72.42 \\
& BlurbsClusteringP2P & 25.06 & 33.77 & 25.28 & 33.09 & 24.04 & 28.95 & 24.10 & 32.63 & 25.62 & 34.85 \\
& BlurbsClusteringS2S & 12.42 & 16.71 & 12.70 & 16.82 & 13.02 & 17.30 & 12.81 & 17.01 & 13.46 & 17.52 \\
& TenKGnadClusteringP2P & 27.51 & 34.42 & 27.39 & 38.47 & 19.15 & 24.18 & 27.45 & 35.49 & 27.33 & 33.44 \\
& TenKGnadClusteringS2S & 19.90 & 29.75 & 18.73 & 28.71 & 18.36 & 30.12 & 21.01 & 30.33 & 22.47 & 26.92 \\
& GermanQuAD-Retrieval & 36.64 & 56.36 & 42.36 & 60.56 & 26.01 & 41.97 & 39.07 & 57.45 & 46.31 & 63.03 \\
& XMarket & 10.58 & 13.51 & 10.05 & 12.57 & 12.51 & 14.44 & 10.89 & 13.96 & 11.28 & 13.60 \\
& GerDaLIR & 0.78 & 1.00 & 0.80 & 1.00 & 0.35 & 0.61 & 0.71 & 1.03 & 1.04 & 1.62 \\
& GermanSTSBenchmark & 53.66 & 66.01 & 56.99 & 68.37 & 50.44 & 62.92 & 57.02 & 68.22 & 60.72 & 68.08 \\
& STS22 & 1.90 & 7.31 & 1.06 & 7.86 & 3.12 & 6.93 & 0.16 & 6.25 & 3.57 & 7.15 \\
\cmidrule(lr){2-12}
& \textbf{Avg.} & 39.40 & \textbf{45.00} & 40.27 & \textbf{45.77} & 38.40 & \textbf{42.55} & 39.94 & \textbf{45.36} & 41.60 & \textbf{45.66} \\
\midrule

% ==================== French ====================
\multirow{24}{*}{fr}
& AmazonReviewsClassification & 40.64 & 41.23 & 41.59 & 41.70 & 40.92 & 40.28 & 40.50 & 41.08 & 41.02 & 41.28 \\
& MasakhaNEWSClassification & 70.88 & 73.67 & 69.79 & 73.36 & 67.91 & 70.31 & 70.76 & 73.58 & 69.60 & 74.69 \\
& MassiveIntentClassification & 64.68 & 67.92 & 66.51 & 68.41 & 66.74 & 67.48 & 65.79 & 68.78 & 67.73 & 68.86 \\
& MassiveScenarioClassification & 68.49 & 71.63 & 69.88 & 71.89 & 70.17 & 70.84 & 69.71 & 72.86 & 71.60 & 72.67 \\
& MTOPDomainClassification & 81.63 & 86.98 & 83.43 & 87.38 & 83.24 & 85.13 & 83.05 & 88.03 & 84.88 & 88.54 \\
& MTOPIntentClassification & 72.87 & 76.00 & 73.91 & 76.40 & 75.21 & 76.62 & 73.65 & 76.62 & 74.12 & 76.77 \\
& AlloProfClusteringS2S & 45.95 & 46.94 & 46.42 & 47.10 & 47.40 & 49.37 & 44.20 & 47.36 & 46.39 & 48.29 \\
& AlloProfClusteringP2P & 52.61 & 56.13 & 51.30 & 55.17 & 54.58 & 54.11 & 51.31 & 55.36 & 53.48 & 56.55 \\
& HALClusteringS2S & 18.86 & 21.83 & 19.59 & 22.16 & 17.25 & 21.44 & 19.34 & 22.47 & 18.61 & 22.30 \\
& MasakhaNEWSClusteringP2P & 31.75 & 54.41 & 30.83 & 48.57 & 27.97 & 45.04 & 31.64 & 55.64 & 21.49 & 49.96 \\
& MasakhaNEWSClusteringS2S & 35.22 & 46.51 & 30.66 & 30.82 & 30.38 & 36.78 & 33.21 & 46.94 & 33.86 & 33.48 \\
& MLSUMClusteringP2P & 31.97 & 35.99 & 29.47 & 34.55 & 31.70 & 32.84 & 32.47 & 37.20 & 32.77 & 36.37 \\
& MLSUMClusteringS2S & 30.76 & 34.67 & 28.23 & 33.26 & 29.54 & 31.60 & 31.29 & 35.37 & 31.99 & 34.87 \\
& AlloprofReranking & 45.73 & 67.27 & 47.82 & 67.33 & 48.13 & 64.32 & 47.15 & 68.21 & 50.97 & 67.67 \\
& SyntecReranking & 45.57 & 78.11 & 54.24 & 77.53 & 55.74 & 74.23 & 49.85 & 79.14 & 66.13 & 79.65 \\
& AlloprofRetrieval & 9.23 & 29.87 & 11.69 & 30.62 & 12.30 & 22.65 & 10.25 & 31.55 & 14.87 & 29.32 \\
& BSARDRetrieval & 27.48 & 41.89 & 34.69 & 48.65 & 20.72 & 38.29 & 30.18 & 44.60 & 37.84 & 47.30 \\
& MintakaRetrieval & 7.95 & 33.71 & 8.89 & 34.38 & 15.88 & 42.77 & 11.47 & 37.21 & 8.49 & 33.08 \\
& SyntecRetrieval & 37.00 & 65.92 & 45.10 & 66.99 & 40.62 & 60.85 & 45.66 & 66.29 & 58.74 & 65.99 \\
& XPQARetrieval & 30.99 & 52.34 & 34.36 & 53.66 & 30.37 & 49.24 & 35.20 & 54.38 & 36.72 & 53.62 \\
& SICKFr & 64.59 & 68.35 & 65.64 & 69.04 & 59.22 & 68.39 & 66.71 & 69.78 & 67.81 & 69.56 \\
& STSBenchmarkMultilingualSTS & 60.98 & 69.67 & 62.75 & 70.96 & 60.62 & 69.79 & 63.75 & 71.33 & 67.39 & 71.56 \\
& STS22 & 40.03 & 43.23 & 36.84 & 41.50 & 47.59 & 53.02 & 43.79 & 46.09 & 38.25 & 45.88 \\
& SummEvalFr & 26.06 & 26.82 & 26.43 & 27.35 & 28.28 & 29.91 & 25.67 & 28.45 & 26.28 & 28.63 \\
\cmidrule(lr){2-12}
& \textbf{Avg.} & 43.41 & \textbf{53.80} & 44.58 & \textbf{53.28} & 44.27 & \textbf{52.30} & 44.86 & \textbf{54.93} & 46.71 & \textbf{54.04} \\
\bottomrule
\end{tabular}
\caption{Detailed monolingual results of training-free baselines with and without \method{} across $17$ non-English languages using LLaMA-2-7B (Part I). For language-level averages, the better result between each baseline and its \method{}-enhanced version is highlighted in \textbf{bold}.}
\label{tab:full-nonenglish-results-1}
\end{table*}

\begin{table*}[t]
\centering
\scriptsize
\setlength{\tabcolsep}{3.2pt}
\begin{tabular}{llcccccccccc}
\toprule
\textbf{Lang.} & \textbf{Dataset}
& \multicolumn{2}{c}{\textbf{PromptEOL}}
& \multicolumn{2}{c}{\textbf{Pretended CoT}}
& \multicolumn{2}{c}{\textbf{ECHO}}
& \multicolumn{2}{c}{\textbf{CP}}
& \multicolumn{2}{c}{\textbf{TP}} \\
\cmidrule(lr){3-4}
\cmidrule(lr){5-6}
\cmidrule(lr){7-8}
\cmidrule(lr){9-10}
\cmidrule(lr){11-12}
& & Base & \textit{w/} \textbf{\method{}}
& Base & \textit{w/} \textbf{\method{}}
& Base & \textit{w/} \textbf{\method{}}
& Base & \textit{w/} \textbf{\method{}}
& Base & \textit{w/} \textbf{\method{}} \\
\midrule

% ==================== Chinese ====================
\multirow{29}{*}{zh}
& CMedQAv1-reranking & 32.64 & 41.45 & 37.17 & 47.32 & 30.67 & 38.29 & 32.63 & 43.70 & 33.07 & 46.81 \\
& CMedQAv2-reranking & 36.68 & 49.75 & 39.76 & 50.65 & 34.88 & 47.12 & 37.04 & 51.06 & 37.07 & 51.28 \\
& MMarcoReranking & 14.19 & 15.01 & 10.07 & 10.89 & 13.30 & 14.45 & 13.05 & 15.24 & 14.16 & 13.68 \\
& T2Reranking & 62.99 & 64.29 & 62.91 & 64.11 & 63.04 & 63.80 & 62.68 & 64.19 & 62.29 & 64.47 \\
& CLSClusteringS2S & 28.75 & 27.98 & 28.56 & 27.74 & 26.28 & 25.79 & 30.18 & 28.45 & 28.01 & 27.50 \\
& CLSClusteringP2P & 27.63 & 28.50 & 29.02 & 29.08 & 28.48 & 28.89 & 29.02 & 29.54 & 30.59 & 32.71 \\
& ThuNewsClusteringS2S & 41.84 & 46.50 & 40.44 & 46.39 & 40.69 & 46.07 & 41.63 & 47.87 & 40.72 & 46.81 \\
& ThuNewsClusteringP2P & 47.64 & 52.53 & 47.61 & 51.91 & 45.72 & 51.60 & 49.01 & 54.69 & 48.24 & 54.40 \\
& LCQMC & 51.87 & 57.32 & 55.85 & 60.62 & 51.86 & 54.85 & 54.13 & 59.16 & 52.59 & 58.43 \\
& PAWSX & 18.67 & 19.48 & 18.73 & 19.07 & 20.39 & 21.19 & 18.83 & 19.34 & 18.32 & 19.78 \\
& QBQTC & 25.83 & 28.54 & 26.91 & 26.85 & 20.83 & 26.43 & 25.98 & 28.18 & 23.64 & 27.36 \\
& Waimai & 83.21 & 82.99 & 83.17 & 82.86 & 83.59 & 81.61 & 83.07 & 82.99 & 83.53 & 83.04 \\
& OnlineShopping & 89.48 & 89.41 & 88.63 & 88.08 & 89.92 & 87.69 & 89.51 & 88.94 & 89.87 & 89.65 \\
& JDReview & 80.36 & 80.53 & 80.56 & 80.17 & 81.26 & 80.96 & 80.68 & 80.56 & 81.78 & 80.62 \\
& MultilingualSentiment & 68.03 & 67.55 & 68.15 & 67.16 & 67.67 & 65.73 & 67.51 & 66.91 & 67.16 & 67.45 \\
& ATEC & 21.54 & 22.61 & 26.73 & 25.25 & 19.49 & 18.40 & 23.67 & 23.18 & 23.53 & 22.56 \\
& BQ & 36.44 & 39.83 & 40.93 & 40.92 & 37.58 & 37.74 & 38.26 & 39.96 & 36.68 & 38.54 \\
& STSB & 60.03 & 66.77 & 65.09 & 69.38 & 59.59 & 63.71 & 62.19 & 68.24 & 58.77 & 66.64 \\
& AFQMC & 16.40 & 17.04 & 21.33 & 19.57 & 15.65 & 14.33 & 17.83 & 17.06 & 17.48 & 16.50 \\
& Cmnli & 62.31 & 63.74 & 64.65 & 64.58 & 63.19 & 62.99 & 62.78 & 63.86 & 62.73 & 63.79 \\
& Ocnli & 55.93 & 56.90 & 56.42 & 57.01 & 56.74 & 57.23 & 56.52 & 56.69 & 57.01 & 57.93 \\
& CmedqaRetrieval & 7.27 & 12.05 & 7.36 & 11.62 & 6.92 & 9.88 & 7.90 & 12.36 & 7.19 & 11.86 \\
& CovidRetrieval & 13.60 & 20.37 & 18.51 & 19.30 & 19.05 & 25.35 & 16.68 & 20.88 & 19.37 & 27.42 \\
& DuRetrieval & 25.85 & 32.89 & 27.21 & 33.71 & 24.84 & 28.89 & 25.02 & 33.17 & 26.26 & 36.68 \\
& EcomRetrieval & 18.25 & 21.17 & 16.86 & 21.04 & 18.65 & 19.74 & 18.45 & 22.11 & 17.41 & 21.22 \\
& MMarcoRetrieval & 32.72 & 39.83 & 35.43 & 41.31 & 33.13 & 37.73 & 35.09 & 40.83 & 35.02 & 42.29 \\
& MedicalRetrieval & 9.19 & 15.03 & 11.33 & 15.82 & 9.35 & 12.99 & 10.21 & 15.32 & 8.96 & 16.39 \\
& T2Retrieval & 24.42 & 30.39 & 25.19 & 30.95 & 25.04 & 28.15 & 23.02 & 30.42 & 23.67 & 32.66 \\
& VideoRetrieval & 27.94 & 31.24 & 25.99 & 29.87 & 26.92 & 29.30 & 27.94 & 31.40 & 29.88 & 32.69 \\
\cmidrule(lr){2-12}
& \textbf{Avg.} & 38.68 & \textbf{42.13} & 40.02 & \textbf{42.53} & 38.44 & \textbf{40.72} & 39.33 & \textbf{42.63} & 39.14 & \textbf{43.14} \\
\midrule

% ==================== Russian ====================
\multirow{18}{*}{ru}
& GeoreviewClassification & 44.47 & 45.09 & 45.30 & 44.43 & 43.49 & 43.34 & 43.65 & 44.39 & 44.91 & 45.84 \\
& HeadlineClassification & 75.76 & 76.57 & 75.23 & 76.18 & 74.04 & 75.62 & 76.19 & 77.12 & 75.23 & 76.38 \\
& InappropriatenessClassification & 59.06 & 60.26 & 59.41 & 60.81 & 57.96 & 58.84 & 58.85 & 59.71 & 59.53 & 60.00 \\
& KinopoiskClassification & 49.41 & 51.22 & 50.09 & 51.08 & 53.11 & 52.17 & 47.94 & 49.99 & 51.54 & 52.77 \\
& MassiveIntentClassification & 66.66 & 69.23 & 67.68 & 69.56 & 67.05 & 68.09 & 67.82 & 70.04 & 68.13 & 69.77 \\
& MassiveScenarioClassification & 69.13 & 73.00 & 69.94 & 73.10 & 69.68 & 71.29 & 70.31 & 73.83 & 70.57 & 73.22 \\
& RuReviewsClassification & 61.64 & 62.05 & 63.12 & 63.15 & 61.86 & 60.54 & 61.31 & 61.80 & 63.42 & 63.54 \\
& RuSciBenchGRNTIClassification & 50.03 & 44.77 & 50.15 & 45.32 & 45.95 & 40.66 & 48.65 & 44.22 & 51.06 & 49.50 \\
& RuSciBenchOECDClassification & 38.17 & 33.77 & 38.49 & 34.86 & 35.24 & 31.07 & 37.21 & 33.39 & 39.44 & 37.33 \\
& GeoreviewClusteringP2P & 36.63 & 47.60 & 38.72 & 51.18 & 30.86 & 40.66 & 35.15 & 45.77 & 33.94 & 44.69 \\
& RuSciBenchGRNTIClusteringP2P & 34.46 & 33.97 & 37.74 & 37.57 & 28.24 & 30.63 & 32.10 & 32.63 & 35.80 & 40.26 \\
& RuSciBenchOECDClusteringP2P & 29.65 & 29.64 & 33.09 & 32.91 & 25.71 & 27.10 & 28.08 & 28.19 & 32.68 & 33.77 \\
& RuBQReranking & 44.51 & 58.68 & 47.55 & 60.42 & 43.47 & 56.11 & 47.79 & 59.49 & 42.17 & 59.84 \\
& MIRACLReranking & 21.85 & 30.86 & 26.98 & 35.01 & 23.29 & 29.60 & 24.22 & 31.89 & 20.86 & 32.47 \\
& RuBQRetrieval & 10.26 & 23.48 & 13.00 & 26.67 & 12.36 & 20.39 & 14.63 & 25.57 & 9.98 & 26.07 \\
& RUParaPhraserSTS & 52.22 & 55.99 & 57.01 & 60.85 & 48.18 & 51.38 & 53.78 & 56.82 & 54.50 & 58.15 \\
& STS22 & 10.10 & 20.27 & 8.74 & 15.16 & 19.16 & 30.62 & 9.80 & 20.39 & 10.35 & 24.59 \\
& RuSTSBenchmarkSTS & 56.25 & 67.04 & 59.13 & 68.39 & 55.59 & 64.55 & 58.56 & 68.55 & 62.02 & 68.60 \\
\cmidrule(lr){2-12}
& \textbf{Avg.} & 45.00 & \textbf{49.08} & 46.74 & \textbf{50.37} & 44.18 & \textbf{47.37} & 45.33 & \textbf{49.10} & 45.90 & \textbf{50.93} \\
\midrule

% ==================== Arabic ====================
\multirow{6}{*}{ar}
& MassiveIntentClassification & 50.72 & 54.33 & 51.11 & 54.11 & 50.85 & 53.78 & 51.14 & 54.52 & 51.53 & 54.35 \\
& MultiHateClassification & 58.75 & 58.86 & 58.47 & 59.76 & 57.43 & 57.91 & 58.85 & 58.83 & 59.21 & 58.56 \\
& MLQARetrieval & 5.92 & 17.42 & 5.01 & 15.98 & 5.31 & 17.00 & 6.88 & 17.43 & 6.16 & 20.69 \\
& XNLI & 59.34 & 61.80 & 60.22 & 61.96 & 60.75 & 60.86 & 59.39 & 61.64 & 59.59 & 62.83 \\
& STS17 & 47.46 & 61.69 & 48.51 & 62.69 & 49.38 & 63.56 & 48.64 & 63.01 & 48.71 & 62.45 \\
& STS22.v2 & 24.70 & 28.55 & 27.88 & 32.62 & 30.80 & 37.65 & 25.38 & 30.03 & 12.64 & 24.31 \\
\cmidrule(lr){2-12}
& \textbf{Avg.} & 41.10 & \textbf{47.11} & 41.87 & \textbf{47.85} & 42.42 & \textbf{48.46} & 41.71 & \textbf{47.58} & 39.64 & \textbf{47.20} \\
\midrule

% ==================== Spanish ====================
\multirow{9}{*}{es}
& CataloniaTweetClassification & 45.93 & 46.65 & 46.46 & 48.48 & 44.65 & 45.30 & 45.63 & 46.59 & 46.13 & 46.81 \\
& MassiveIntentClassification & 64.69 & 67.46 & 66.53 & 68.03 & 67.04 & 67.24 & 65.59 & 68.16 & 66.78 & 67.98 \\
& MultiHateClassification & 65.06 & 66.81 & 65.24 & 67.55 & 67.58 & 66.36 & 64.94 & 66.71 & 66.85 & 65.87 \\
& SIB200ClusteringS2S & 25.49 & 43.49 & 25.14 & 43.12 & 28.44 & 41.17 & 26.78 & 43.89 & 29.60 & 41.82 \\
& BelebeleRetrieval & 36.16 & 68.52 & 39.85 & 71.03 & 28.94 & 65.38 & 45.26 & 71.57 & 40.08 & 76.25 \\
& MLQARetrieval & 17.57 & 35.62 & 23.26 & 39.14 & 18.68 & 37.97 & 23.18 & 37.00 & 21.88 & 42.00 \\
& XNLI & 61.57 & 64.49 & 64.08 & 66.88 & 63.05 & 63.96 & 62.05 & 64.49 & 64.58 & 66.29 \\
& STS17 & 71.51 & 77.74 & 74.19 & 80.69 & 68.56 & 78.33 & 73.06 & 79.63 & 77.83 & 80.39 \\
& STS22.v2 & 43.49 & 48.17 & 40.93 & 47.19 & 47.88 & 52.48 & 44.81 & 49.75 & 47.53 & 54.01 \\
\cmidrule(lr){2-12}
& \textbf{Avg.} & 47.90 & \textbf{57.66} & 49.52 & \textbf{59.12} & 48.31 & \textbf{57.57} & 50.15 & \textbf{58.64} & 51.25 & \textbf{60.16} \\
\bottomrule
\end{tabular}
\caption{Detailed monolingual results of training-free baselines with and without \method{} across $17$ non-English languages using LLaMA-2-7B (Part II). For language-level averages, the better result between each baseline and its \method{}-enhanced version is highlighted in \textbf{bold}.}
\label{tab:full-nonenglish-results-2}
\end{table*}

\begin{table*}[t]
\centering
\scriptsize
\setlength{\tabcolsep}{3.2pt}
\begin{tabular}{llcccccccccc}
\toprule
\textbf{Lang.} & \textbf{Dataset}
& \multicolumn{2}{c}{\textbf{PromptEOL}}
& \multicolumn{2}{c}{\textbf{Pretended CoT}}
& \multicolumn{2}{c}{\textbf{ECHO}}
& \multicolumn{2}{c}{\textbf{CP}}
& \multicolumn{2}{c}{\textbf{TP}} \\
\cmidrule(lr){3-4}
\cmidrule(lr){5-6}
\cmidrule(lr){7-8}
\cmidrule(lr){9-10}
\cmidrule(lr){11-12}
& & Base & \textit{w/} \textbf{\method{}}
& Base & \textit{w/} \textbf{\method{}}
& Base & \textit{w/} \textbf{\method{}}
& Base & \textit{w/} \textbf{\method{}}
& Base & \textit{w/} \textbf{\method{}} \\
\midrule

% ==================== Swahili ====================
\multirow{5}{*}{sw}
& AfriSentiClassification & 36.35 & 36.38 & 35.94 & 36.63 & 38.57 & 36.74 & 36.60 & 36.82 & 37.49 & 37.34 \\
& MasakhaNEWSClassification & 66.76 & 69.12 & 66.11 & 68.24 & 63.66 & 65.27 & 67.35 & 69.60 & 66.85 & 68.68 \\
& MassiveIntentClassification & 37.03 & 43.30 & 38.00 & 42.53 & 37.19 & 40.91 & 36.79 & 43.28 & 38.58 & 44.12 \\
& MasakhaNEWSClusteringS2S & 20.78 & 22.79 & 11.73 & 24.53 & 17.08 & 21.91 & 19.64 & 22.29 & 20.00 & 23.32 \\
& XNLI & 52.93 & 56.67 & 52.77 & 56.82 & 52.85 & 54.54 & 53.02 & 56.67 & 53.32 & 56.79 \\
\cmidrule(lr){2-12}
& \textbf{Avg.} & 42.80 & \textbf{45.65} & 40.91 & \textbf{45.75} & 41.87 & \textbf{43.87} & 42.68 & \textbf{45.73} & 43.25 & \textbf{46.05} \\
\midrule

% ==================== Thai ====================
\multirow{4}{*}{th}
& MassiveIntentClassification & 52.10 & 54.36 & 52.84 & 54.80 & 51.88 & 53.57 & 52.41 & 54.86 & 52.64 & 54.54 \\
& SIB200ClusteringS2S & 23.03 & 41.50 & 24.92 & 38.88 & 22.10 & 36.79 & 23.80 & 38.44 & 23.22 & 36.08 \\
& XNLI & 59.77 & 61.76 & 60.75 & 62.55 & 61.29 & 61.59 & 59.96 & 61.86 & 59.80 & 62.83 \\
& BelebeleRetrieval & 28.48 & 41.06 & 29.84 & 43.72 & 24.17 & 35.65 & 28.86 & 41.96 & 25.89 & 41.48 \\
\cmidrule(lr){2-12}
& \textbf{Avg.} & 40.80 & \textbf{49.67} & 42.09 & \textbf{49.99} & 39.86 & \textbf{46.90} & 41.26 & \textbf{49.28} & 40.39 & \textbf{48.73} \\
\midrule

% ==================== Indonesian ====================
\multirow{6}{*}{id}
& MassiveIntentClassification & 62.84 & 65.13 & 64.21 & 65.72 & 62.97 & 64.35 & 63.58 & 65.73 & 64.59 & 65.76 \\
& NusaX-senti & 82.78 & 79.45 & 82.13 & 78.03 & 81.45 & 77.40 & 82.20 & 79.10 & 84.78 & 80.35 \\
& SIB200ClusteringS2S & 26.50 & 44.60 & 25.72 & 41.47 & 27.13 & 40.99 & 25.97 & 41.96 & 28.23 & 43.50 \\
& indonli & 50.23 & 50.68 & 50.47 & 50.98 & 48.82 & 49.23 & 50.25 & 50.64 & 51.55 & 51.60 \\
& BelebeleRetrieval & 24.50 & 64.69 & 42.19 & 66.11 & 23.93 & 49.31 & 45.97 & 65.81 & 41.35 & 69.89 \\
& SemRel24STS & 50.91 & 51.24 & 53.24 & 52.73 & 49.97 & 50.61 & 51.06 & 51.36 & 52.19 & 52.43 \\
\cmidrule(lr){2-12}
& \textbf{Avg.} & 49.60 & \textbf{59.30} & 52.99 & \textbf{59.17} & 49.04 & \textbf{55.31} & 53.17 & \textbf{59.10} & 53.78 & \textbf{60.59} \\
\midrule

% ==================== Italian ====================
\multirow{10}{*}{it}
& ItaCaseholdClassification & 86.56 & 87.06 & 86.52 & 84.93 & 89.32 & 89.55 & 86.79 & 86.24 & 84.25 & 86.11 \\
& MassiveIntentClassification & 64.97 & 67.91 & 67.03 & 68.79 & 68.35 & 68.04 & 66.13 & 68.86 & 67.95 & 69.13 \\
& MultiHateClassification & 61.27 & 63.59 & 62.17 & 63.69 & 63.67 & 63.36 & 61.55 & 63.70 & 62.93 & 63.08 \\
& SwissJudgementClassification & 54.08 & 54.99 & 53.85 & 55.57 & 55.95 & 58.14 & 53.57 & 55.32 & 52.03 & 51.87 \\
& SIB200ClusteringS2S & 29.49 & 41.24 & 29.00 & 40.72 & 30.26 & 42.18 & 28.26 & 42.39 & 31.00 & 41.18 \\
& BelebeleRetrieval & 37.97 & 72.12 & 46.73 & 75.43 & 30.17 & 64.50 & 48.07 & 75.14 & 44.96 & 74.73 \\
& WikipediaRetrievalMultilingual & 30.86 & 58.85 & 44.05 & 66.40 & 29.07 & 62.46 & 41.72 & 62.76 & 37.99 & 64.90 \\
& RTE3 & 85.78 & 86.44 & 85.57 & 86.51 & 86.20 & 86.87 & 85.79 & 86.55 & 86.84 & 87.04 \\
& WikipediaRerankingMultilingual & 62.27 & 80.12 & 68.98 & 83.00 & 63.29 & 80.75 & 67.22 & 80.98 & 65.33 & 80.87 \\
& STS22.v2 & 38.12 & 47.83 & 41.58 & 49.98 & 49.04 & 53.63 & 39.94 & 49.76 & 41.03 & 52.90 \\
\cmidrule(lr){2-12}
& \textbf{Avg.} & 55.10 & \textbf{66.01} & 58.55 & \textbf{67.50} & 56.53 & \textbf{66.95} & 57.90 & \textbf{67.17} & 57.43 & \textbf{67.18} \\
\midrule

% ==================== Malay ====================
\multirow{2}{*}{ms}
& SIB200ClusteringS2S & 25.56 & 38.52 & 23.63 & 40.66 & 22.67 & 39.63 & 23.72 & 41.24 & 26.23 & 40.76 \\
& BelebeleRetrieval & 24.79 & 64.50 & 44.36 & 64.78 & 20.25 & 42.86 & 48.26 & 64.79 & 39.64 & 69.76 \\
\cmidrule(lr){2-12}
& \textbf{Avg.} & 25.20 & \textbf{51.51} & 34.00 & \textbf{52.72} & 21.46 & \textbf{41.24} & 35.99 & \textbf{53.02} & 32.93 & \textbf{55.26} \\
\midrule

% ==================== Portuguese ====================
\multirow{7}{*}{pt}
& AfriSentiClassification & 42.56 & 42.69 & 42.78 & 43.77 & 41.95 & 39.93 & 43.01 & 42.69 & 42.35 & 42.12 \\
& MassiveIntentClassification & 64.20 & 67.13 & 66.16 & 67.80 & 66.55 & 66.67 & 65.13 & 68.05 & 66.70 & 68.08 \\
& MultiHateClassification & 65.70 & 66.80 & 65.76 & 66.97 & 65.31 & 65.73 & 65.94 & 66.16 & 65.70 & 65.37 \\
& SIB200ClusteringS2S & 26.33 & 44.94 & 26.06 & 40.04 & 27.71 & 42.66 & 27.61 & 43.40 & 30.66 & 38.19 \\
& BelebeleRetrieval & 37.64 & 70.46 & 46.53 & 73.59 & 31.27 & 66.64 & 43.36 & 73.09 & 41.97 & 74.86 \\
& WikipediaRetrievalMultilingual & 34.82 & 62.62 & 46.50 & 69.35 & 33.91 & 63.20 & 44.47 & 65.40 & 41.93 & 67.89 \\
& WikipediaRerankingMultilingual & 63.78 & 79.15 & 69.18 & 82.00 & 63.03 & 79.17 & 67.07 & 80.12 & 67.08 & 80.93 \\
\cmidrule(lr){2-12}
& \textbf{Avg.} & 47.90 & \textbf{61.97} & 51.85 & \textbf{63.36} & 47.10 & \textbf{60.57} & 50.94 & \textbf{62.70} & 50.91 & \textbf{62.49} \\
\midrule

% ==================== Telugu ====================
\multirow{4}{*}{te}
& MassiveIntentClassification & 28.23 & 37.39 & 29.50 & 38.10 & 28.26 & 34.90 & 28.35 & 37.91 & 28.95 & 39.44 \\
& SIB200ClusteringS2S & 5.57 & 7.54 & 5.16 & 7.21 & 4.75 & 7.06 & 5.46 & 7.28 & 5.06 & 7.87 \\
& BelebeleRetrieval & 7.61 & 15.08 & 7.79 & 14.45 & 7.66 & 13.33 & 7.75 & 16.97 & 6.96 & 16.43 \\
& SemRel24STS & 32.06 & 55.24 & 36.10 & 52.62 & 42.15 & 39.39 & 31.83 & 57.55 & 31.72 & 53.54 \\
\cmidrule(lr){2-12}
% PromptEOL Base Avg.: 18.37 is provisional, computed from the four displayed
% Base scores (28.23, 5.57, 7.61, 32.06); confirm using the original raw log.
& \textbf{Avg.} & 18.37 & \textbf{28.81} & 19.64 & \textbf{28.10} & 20.70 & \textbf{23.67} & 18.35 & \textbf{29.93} & 18.17 & \textbf{29.32} \\
\midrule

% ==================== Vietnamese ====================
\multirow{5}{*}{vi}
& MassiveIntentClassification & 57.70 & 61.03 & 58.65 & 61.40 & 58.36 & 60.29 & 58.37 & 61.89 & 59.14 & 61.46 \\
& SIB200ClusteringS2S & 26.28 & 42.94 & 27.56 & 44.55 & 26.82 & 42.36 & 27.31 & 44.79 & 31.39 & 45.50 \\
& BelebeleRetrieval & 50.01 & 68.11 & 52.37 & 66.87 & 36.03 & 57.90 & 52.57 & 68.96 & 52.11 & 72.08 \\
& MLQARetrieval & 26.78 & 35.55 & 27.69 & 35.36 & 18.39 & 31.24 & 28.26 & 35.35 & 28.02 & 38.39 \\
& XNLI & 62.70 & 65.04 & 63.52 & 65.33 & 62.70 & 63.91 & 63.23 & 64.75 & 65.01 & 66.46 \\
\cmidrule(lr){2-12}
& \textbf{Avg.} & 44.70 & \textbf{54.53} & 45.96 & \textbf{54.70} & 40.46 & \textbf{51.14} & 45.95 & \textbf{55.15} & 47.13 & \textbf{56.78} \\
\midrule

% ==================== Korean ====================
\multirow{5}{*}{ko}
& KorSTS & 56.21 & 66.89 & 58.05 & 67.74 & 54.02 & 64.29 & 59.25 & 68.94 & 58.62 & 67.10 \\
& KLUE-STS & 64.93 & 70.94 & 64.60 & 70.10 & 66.05 & 72.79 & 66.20 & 72.21 & 65.43 & 69.75 \\
& KLUE-TC & 57.77 & 56.20 & 56.25 & 55.53 & 60.65 & 60.75 & 57.35 & 56.53 & 57.50 & 55.54 \\
& MIRACLReranking & 51.90 & 56.85 & 33.52 & 40.59 & 36.00 & 32.22 & 39.41 & 57.20 & 31.10 & 55.77 \\
& Ko-StrategyQA & 34.39 & 50.14 & 33.98 & 49.03 & 20.55 & 45.73 & 36.69 & 48.70 & 26.13 & 47.92 \\
\cmidrule(lr){2-12}
& \textbf{Avg.} & 53.00 & \textbf{60.21} & 49.28 & \textbf{56.60} & 47.46 & \textbf{55.15} & 51.78 & \textbf{60.72} & 47.75 & \textbf{59.22} \\
\midrule

% ==================== Bengali ====================
\multirow{5}{*}{bn}
& MassiveIntentClassification & 42.69 & 47.17 & 43.38 & 47.78 & 44.11 & 47.49 & 42.74 & 47.68 & 43.50 & 48.38 \\
& SIB200ClusteringS2S & 10.92 & 17.65 & 11.44 & 18.92 & 11.47 & 18.48 & 9.78 & 19.29 & 11.74 & 17.19 \\
& WikipediaRerankingMultilingual & 62.96 & 66.80 & 63.38 & 69.19 & 51.70 & 64.35 & 63.26 & 66.99 & 61.17 & 67.41 \\
& BelebeleRetrieval & 15.72 & 20.92 & 15.96 & 26.59 & 9.61 & 21.92 & 15.87 & 23.32 & 14.73 & 27.87 \\
& WikipediaRetrievalMultilingual & 27.76 & 29.63 & 26.89 & 33.70 & 12.14 & 27.52 & 28.73 & 31.00 & 25.32 & 32.11 \\
\cmidrule(lr){2-12}
& \textbf{Avg.} & 32.00 & \textbf{36.43} & 32.21 & \textbf{39.24} & 25.81 & \textbf{35.95} & 32.08 & \textbf{37.66} & 31.29 & \textbf{38.59} \\

\bottomrule
\end{tabular}
\caption{Detailed monolingual results of training-free baselines with and without \method{} across $17$ non-English languages using LLaMA-2-7B (Part III). For language-level averages, the better result between each baseline and its \method{}-enhanced version is highlighted in \textbf{bold}.}
\label{tab:full-nonenglish-results-3}
\end{table*}

\end{document}